\documentclass[runningheads]{llncs}
\usepackage[T1]{fontenc}
\usepackage{graphicx}
\usepackage{hyperref}
\usepackage{color}

\usepackage[capitalize]{cleveref}
\usepackage{caption}
\usepackage{subcaption}
\usepackage{algpseudocode}
\usepackage{algorithm}
\begin{document}
\title{Object Pose Estimation Annotation Pipeline for Multi-view Monocular Camera Systems in Industrial Settings}

%
\titlerunning{Object Pose Estimation Annotation Pipeline}
%
\author{Hazem Youssef\orcidID{0000-0002-7197-9127} \and \\ Frederik Polachowski\orcidID{0000-0002-4470-6270} \and \\ Jérôme Rutinowski\orcidID{0000-0001-6907-9296} \and \\ Moritz Roidl\orcidID{0000-0001-7551-9163} \and \\ Christopher Reining\orcidID{0000-0003-4915-4070}
}
\authorrunning{H. Youssef et al.}
%
\institute{TU Dortmund University, Dortmund, Germany \\ 
Corresponding author: \email{jerome.rutinowski@tu-dortmund.de}
}
\maketitle              
\begin{abstract}
Object localization, and more specifically object pose estimation, in large industrial spaces such as warehouses and production facilities, is essential for material flow operations. Traditional approaches rely on artificial artifacts installed in the environment or excessively expensive equipment, that is not suitable at scale. A more practical approach is to utilize existing cameras in such spaces in order to address the underlying pose estimation problem and to localize objects of interest. In order to leverage state-of-the-art methods in deep learning for object pose estimation, large amounts of data need to be collected and annotated. In this work, we provide an approach to the annotation of large datasets of monocular images without the need for manual labor. Our approach localizes cameras in space, unifies their location with a motion capture system, and uses a set of linear mappings to project 3D models of objects of interest at their ground truth 6D pose locations. We test our pipeline on a custom dataset collected from a system of eight cameras in an industrial setting that mimics the intended area of operation. 
Our approach was able to provide consistent quality annotations for our dataset with $26,482$ object instances at a fraction of the time required by human annotators.

\end{abstract}
\section{Introduction} \label{sec:intro}
6D object pose estimation is the task of determining the spatial pose (i.e., the position and orientation) of a subject of interest along six degrees of freedom, namely along the three translational and three rotational axes in space \cite{hodan_evaluation_2016}.
This task is commonly encountered in the field of robotics \cite{du_vision-based_2021,zeng_multi-view_2017}, when grasping, handling, or localizing objects, which is enabled and facilitated by a successful a priori estimation of the 6D pose of the object in question.
To visually estimate the pose of an object, multiple approaches can be taken.
While using a single sensor might be enough, using more than one might be beneficial as to achieve a higher pose estimation accuracy.
As such, many industrial environments already provide the necessary circumstances for a multi-camera approach, e.g., when exploiting pre-existing infrastructure like surveillance cameras or the footage of cameras that AGVs might use for navigation purposes.
However, when using multiple cameras, the amount of footage that needs to be annotated increases as well.
Even while using a single camera, manual annotation can be cumbersome and inaccurate \cite{russakovsky2015crowdsourcing}.
Using more than one view can therefore seem unfeasible due to annotation and pre-processing overhead.
In addition, especially for industrial applications, pre-annotated datasets are rare to encounter and can therefore seldom be used to train or test a newly developed pose estimation model.
To mitigate such issues, we propose a pipeline to annotate monocular images in a fully automated fashion. The pipeline generates bounding box and mask annotations using the projection of 3D object models at their relative poses, as obtained from the real scene. We also provide a newly collected multi-view dataset as proof of concept of our pipeline. 
The contributions of this work are summarized as follows:
\begin{itemize}
    \item An automated annotation pipeline that outputs camera-relative 6D object poses and bounding boxes from multi-camera input streams
    \item A camera localization method for large indoor spaces
    \item A novel dataset for object pose estimation in industrial-like settings
\end{itemize}

\section{Related Work} \label{sec:related_work}

We first review the existing approaches addressing the 6D object pose estimation task in single-view as well as multi-view settings. We then discuss relevant multi-view datasets and extend our scope to datasets collected in industrial settings. Finally, we discuss existing attempts in the relevant literature to automatically annotate camera input streams.

In recent years, several deep learning-based approaches have been devised for the task of object pose estimation. Such approaches differ in several aspects, including the type of input stream, the number of scene perspectives considered, and the underlying processing stages. In terms of methodology, approaches include template-matching methods such as \cite{park_multi-task_2019,nguyen_templates_2022} that rely on a pre-created set of templates for each object that is associated with ground truth poses and matched to scene objects. 
Feature-based methods, on the other hand, rely on the extraction and matching of special features such as point-pair features \cite{vidal_6d_2018} or 3D local features \cite{buch_local_2016}. Other methods try to learn the pose of scene objects directly from monocular input images \cite{xiang_posecnn_2018} or from RGB-D data of the scene \cite{wang2019densefusion} using end-to-end deep learning architectures. Multi-view and learning-based object pose estimation approaches in particular have achieved significant performance outcomes in recent years, such as \cite{labbe2020,shugurov_multi-view_2021}

For supervised deep learning object pose estimation methods, large amounts of training data are a prerequisite. Since most of the object pose estimation approaches target grasping applications as in \cite{marwan_comprehensive_2021}, the objects found in common benchmark datasets are either for household or toy-like objects \cite{lin2021fusion,hodan_t-less_2017,xiang_posecnn_2018}. Very few datasets target logistics applications or objects commonly found in industrial settings. Although the T-less dataset \cite{hodan_t-less_2017} includes industrial objects, the objects are small-scale and are most similar to those encountered in bin-picking scenarios. This is very different, in terms of setting, from datasets for large-scale localization, using pose estimation as it is done in our work.
The datasets that are most similar to ours, in terms of the target application, are \cite{mayershofer_loco_2020,akar_synthetic_2022,ahmadyan_objectron_2021}. The LOCO dataset \cite{mayershofer_loco_2020} contains large industrial objects recorded in logistics settings. However, the dataset only targets the problem of object detection and thus does not contain pose information for objects of interest. Another dataset that includes slightly larger objects, in comparison to the commonly used household objects in pose estimation settings, is the Objectron dataset \cite{ahmadyan_objectron_2021}. This dataset contains objects such as chairs, bags, and bikes. The dataset is again only concerned with the task of object detection with a focus on outdoor settings. The BMW dataset \cite{akar_synthetic_2022}, on the other hand, is geared towards indoor logistics settings with full-scale industrial objects. However, the dataset is synthetic in nature, with photo-realistic data that targets tasks such as classification, object detection, and segmentation. Other datasets resemble ours in terms of the system layout. In particular, one dataset we were able to encounter, which is publicly available, is BigBird \cite{singh_bigbird_2014}. The dataset is captured through a stationary monocular camera system that offers five perspectives of the scene. The dataset, however, includes only household objects, which are different in scale, form, and texture from industrial objects. 

These datasets vary considerably in size, with datasets ranging from several hundreds to hundreds of thousands of images. Such a volume of data is usually manually annotated to train machine learning architectures. The effort is significantly amplified when considering the full pose annotation needed. Only very few approaches in the literature try to mitigate the problem by offering annotation pipelines with some degree of automation. The approaches in \cite{leibe_objectnet3d_2016,xiang_beyond_2014} offer annotation tools that facilitate the annotation task. They rely on the keypoint matching between a projected object model and the input images. The matching process itself is performed manually by human users, where for each input image the user has to choose the corresponding object model along with matching unique keypoints such as corners, blobs, etc. between the two. The approaches target object detection and object recognition tasks and do not offer a technique to retrieve the pose of the object after keypoint matching is performed.


\section{Dataset Recording} \label{sec:dataset}

\begin{figure}[h]
    \centering
    \includegraphics[width=1\linewidth]{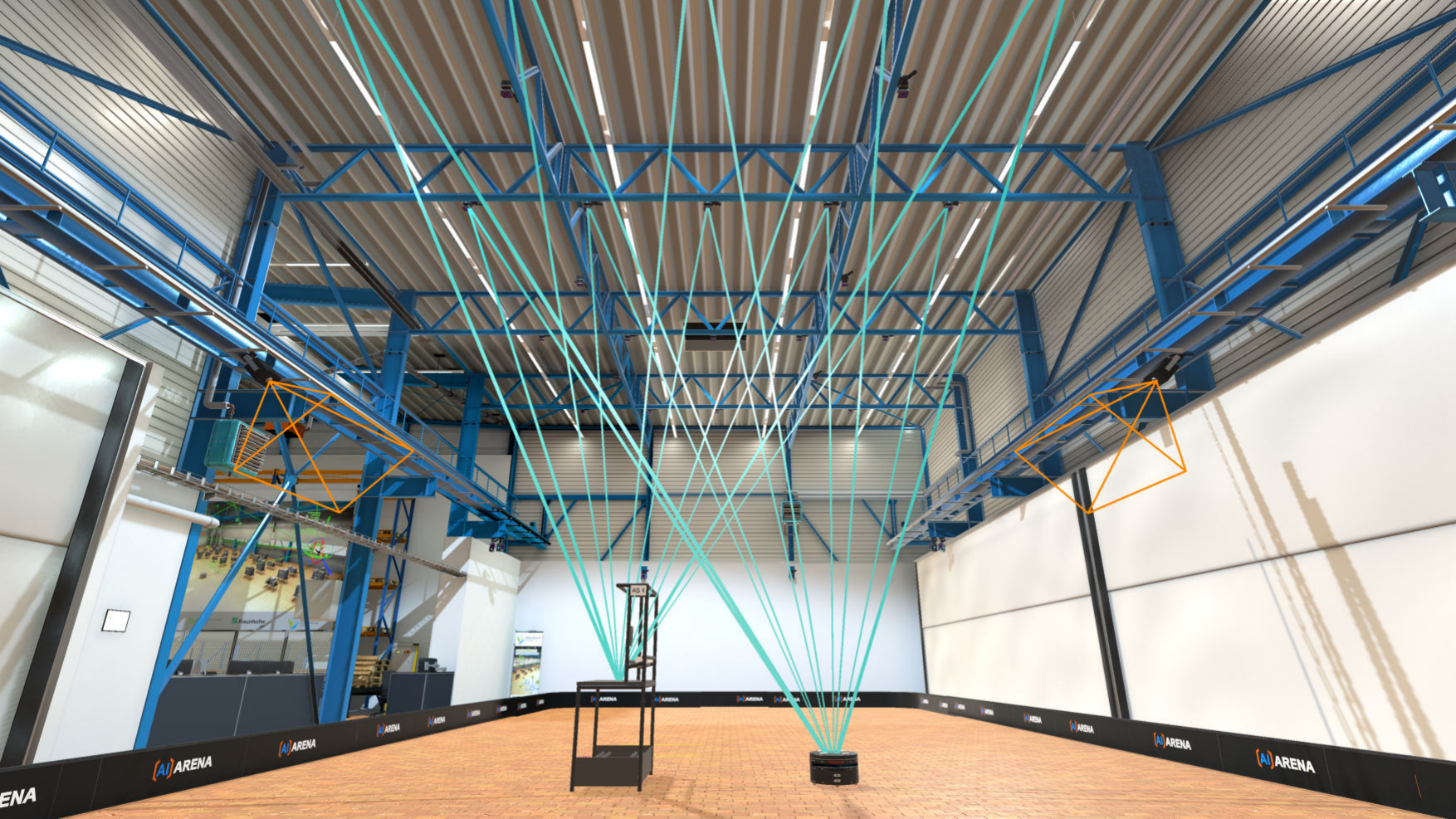}
    \caption{Rendering of our research facility, depicting the camera system used for our recordings. Enlarged representations of two RGB cameras are shown in orange. Reflected rays captured via the motion capture system are shown in cyan.}
    \label{fig:cam_sys}
\end{figure}

Due to the previously mentioned limitations in object pose estimation datasets, we contribute a novel dataset that is collected in an industrial-like environment. We call the dataset \emph{Multi-log}, in reference to multi-view logistics.
Multi-log is an industrial dataset that targets logistics scenarios in which large objects in indoor settings are of interest.
The dataset offers a unique combination of wide-angle monocular RGB images, that are automatically annotated, as discussed in \cref{sec:pipeline}. The dataset was recorded in a small warehouse-like setup, in which eight monocular RGB cameras are installed as shown in \cref{fig:cam_sys}. The cameras are of type Genie Nano C2590 that are capable of capturing 2 MP images. The distance of an object in the area to any of the cameras exceeds $6$ m, which is a major difference between our dataset and existing ones. The area is also covered with $52$ motion capture cameras that offer accurate poses of the objects, with sub-millimeter precision. The acquired poses are used in the automated annotation process of the objects moving in the scene. 

The dataset recording area resembles a small-scale controlled logistics environment. The recording process is performed by deploying the objects in the recording area and moving them, both randomly and in a pre-determined manner. The movement of tracked objects in such area is captured by the motion capture system and the RGB camera system. Using both systems during the capturing process provides continuous image streams from all eight cameras and, simultaneously, the ground truth 6D pose of the objects in each frame. Objects are moved around using a manually controlled mobile robot. 

\begin{figure}[h]
    \centering
    \begin{subfigure}[b]{0.23\textwidth}
        \includegraphics[width=\textwidth]{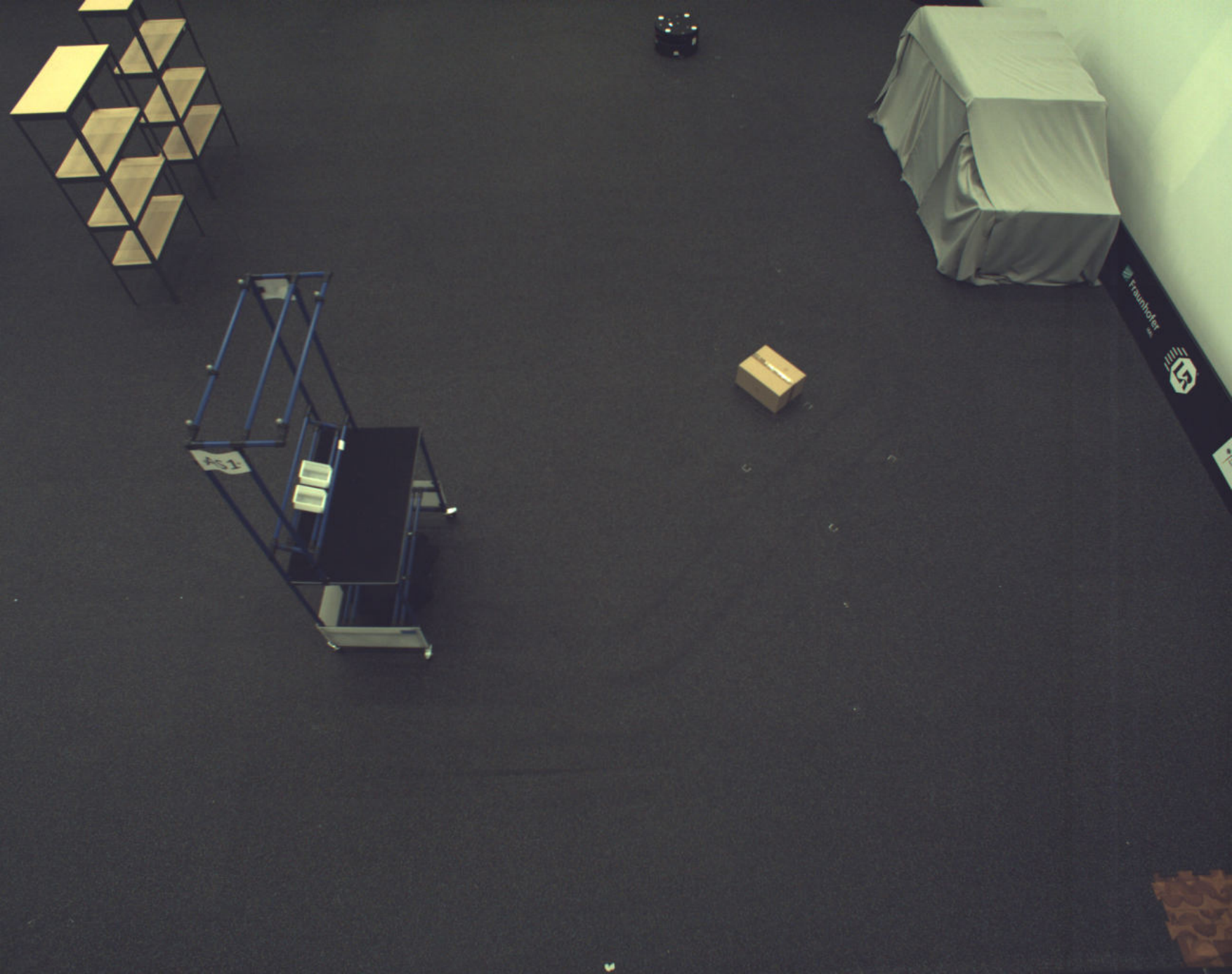}
        \caption{}
        \label{fig:test_cam0}
    \end{subfigure}
    \hfill
    \begin{subfigure}[b]{0.23\textwidth}
        \includegraphics[width=\textwidth]{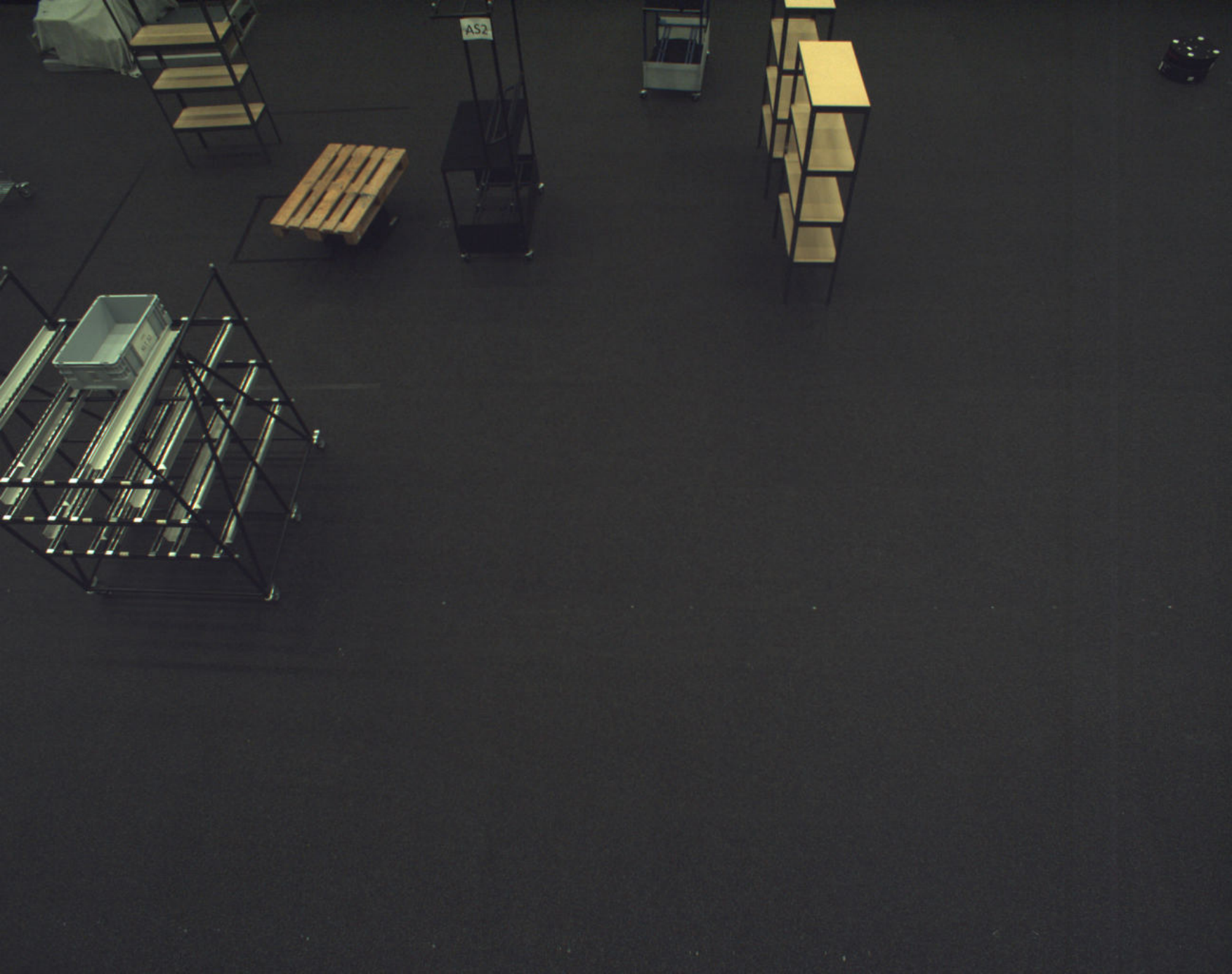}
        \caption{}
        \label{fig:test_cam1}
    \end{subfigure}
    \hfill
    \begin{subfigure}[b]{0.23\textwidth}
        \includegraphics[width=\textwidth]{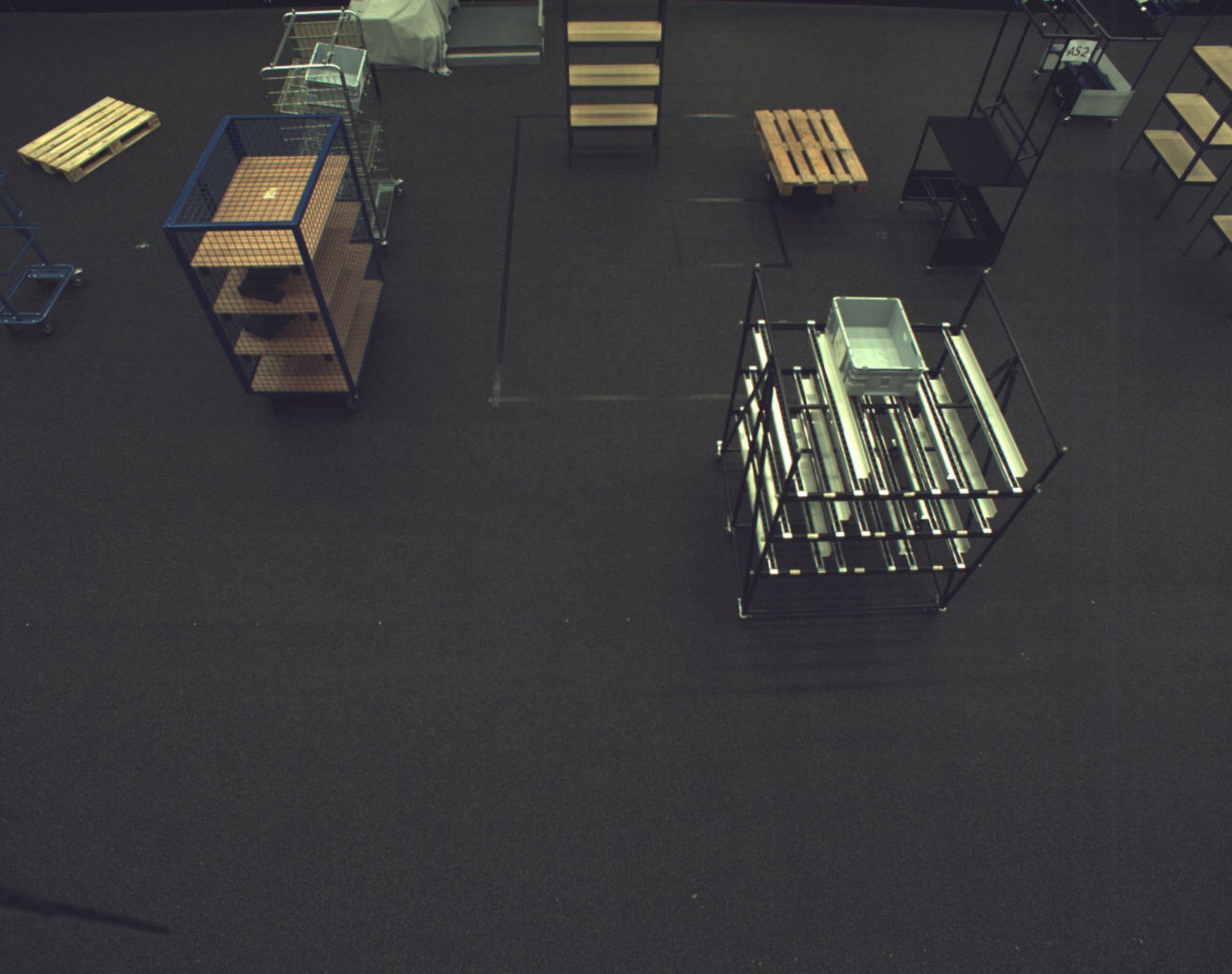}
        \caption{}
        \label{fig:test_cam2}
    \end{subfigure}
    \hfill
    \begin{subfigure}[b]{0.23\textwidth}
        \includegraphics[width=\textwidth]{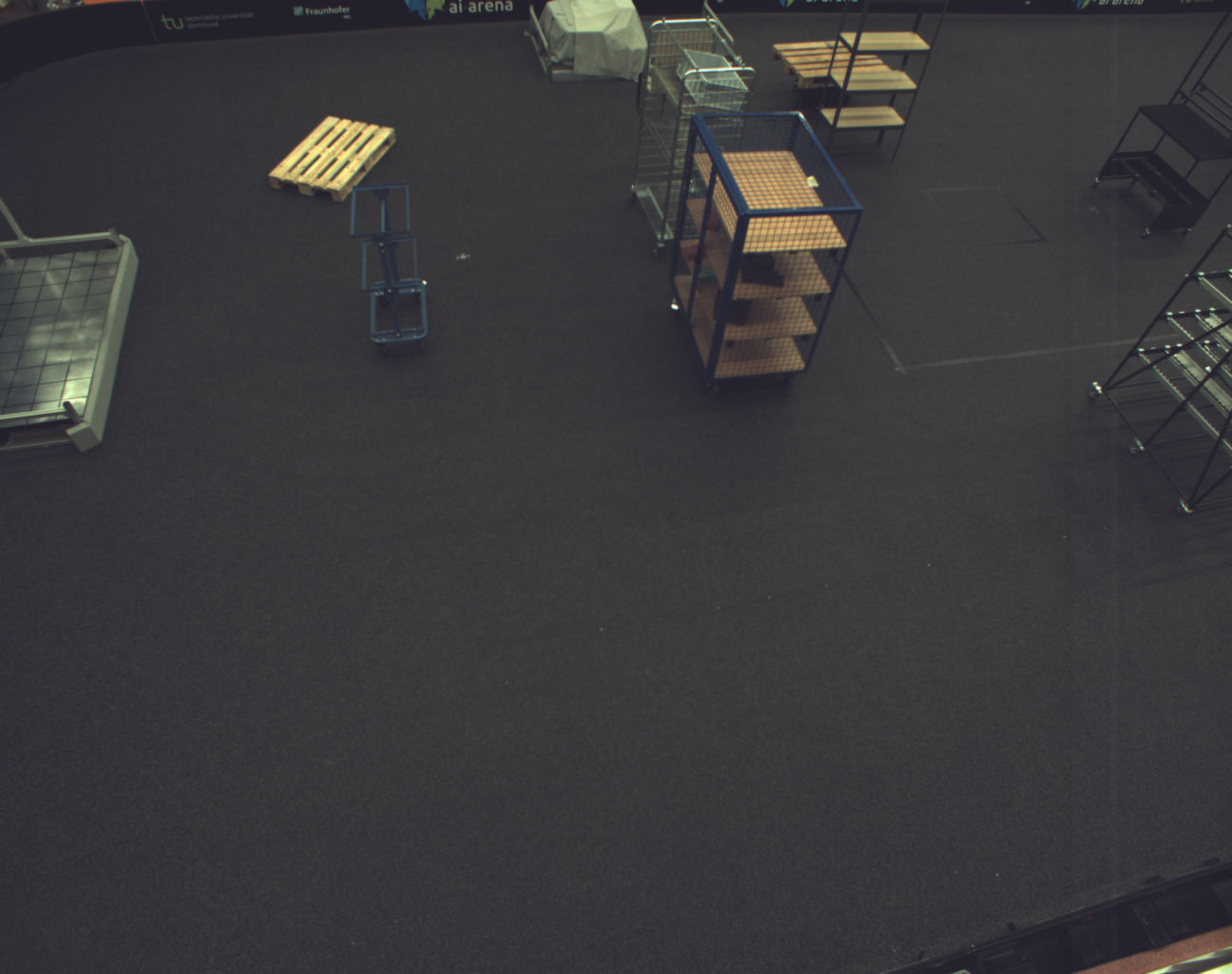}
        \caption{}
        \label{fig:test_cam3}
    \end{subfigure}
        \hfill
    \begin{subfigure}[b]{0.23\textwidth}
        \includegraphics[width=\textwidth]{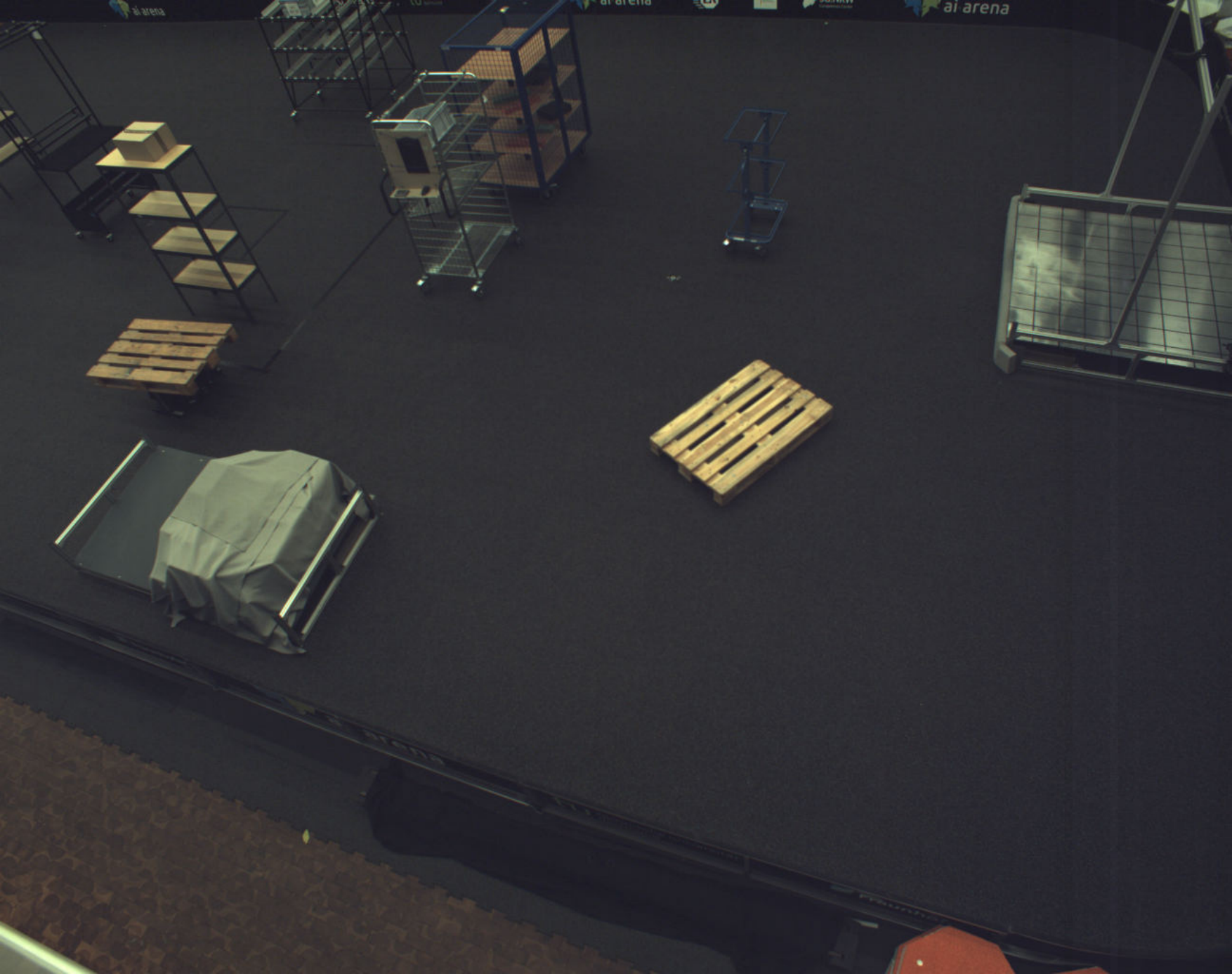}
        \caption{}
        \label{fig:test_cam4}
    \end{subfigure}
    \hfill
    \begin{subfigure}[b]{0.23\textwidth}
        \includegraphics[width=\textwidth]{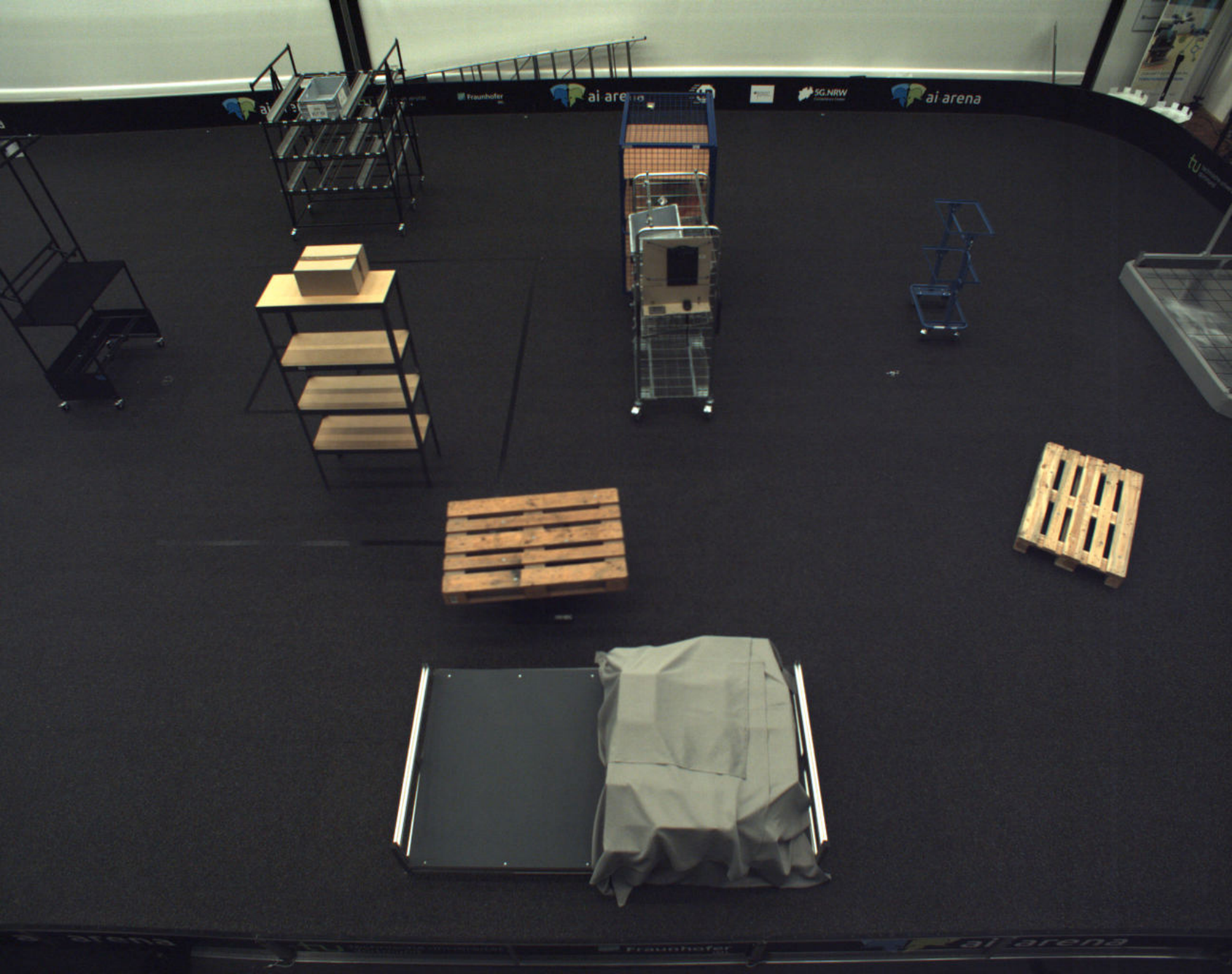}
        \caption{}
        \label{fig:test_cam5}
    \end{subfigure}
    \hfill
    \begin{subfigure}[b]{0.23\textwidth}
        \includegraphics[width=\textwidth]{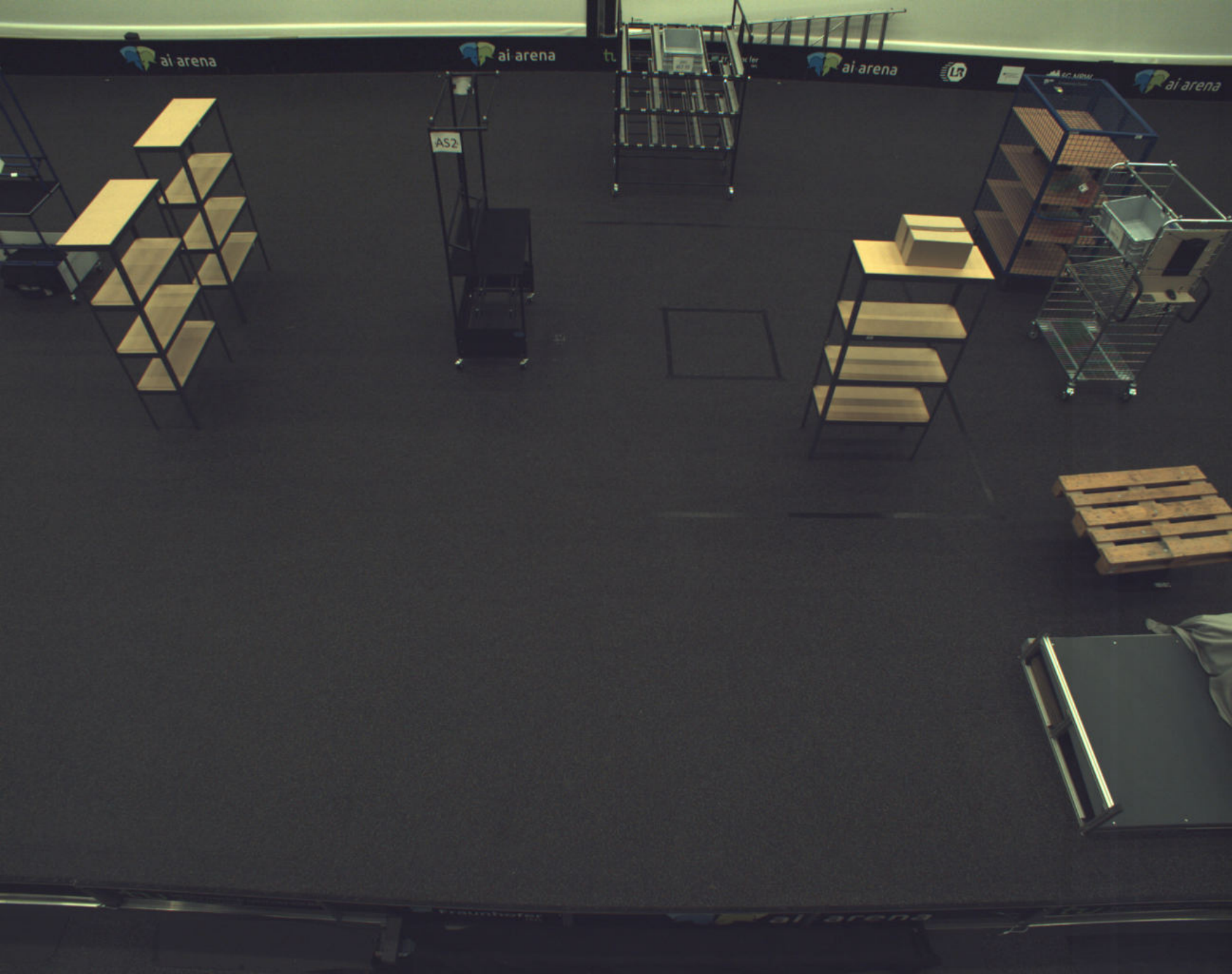}
        \caption{}
        \label{fig:test_cam6}
    \end{subfigure}
    \hfill
    \begin{subfigure}[b]{0.23\textwidth}
        \includegraphics[width=\textwidth]{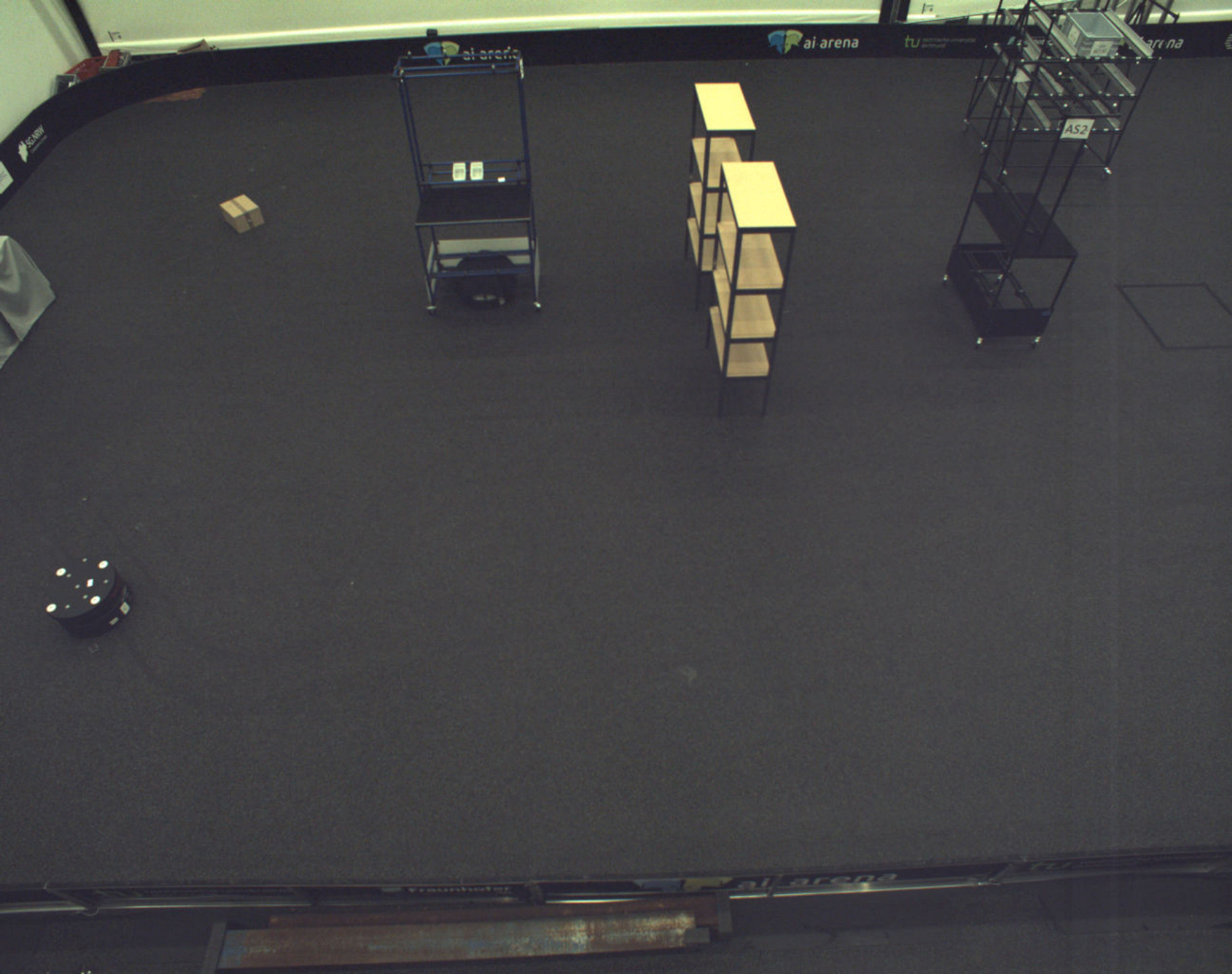}
        \caption{}
        \label{fig:test_cam7}
    \end{subfigure}
    \caption{Sample scene from our dataset as viewed by all eight RGB cameras.}
    \label{fig:testing_samples}
\end{figure}

The environment layout during the collection process is set up in such a way that it mimics a dynamic production facility. The main aim of such a setup is to reduce fingerprinting effects in image detection and segmentation methods that could be caused by the mostly neutral background. The dataset is collected in two different setups that differ in the layout and the stationary untracked objects. Untracked objects used include shelves, roller racks, and commissioning wagons. During any given recording, two to three objects are moving simultaneously. All objects were used during each collection run but were positioned differently. Samples of the test set in one setting from all eight cameras are shown in \cref{fig:testing_samples}.

\begin{figure}[t!]
    \centering
    \begin{subfigure}[b]{0.19\textwidth}
        \includegraphics[width=\textwidth]{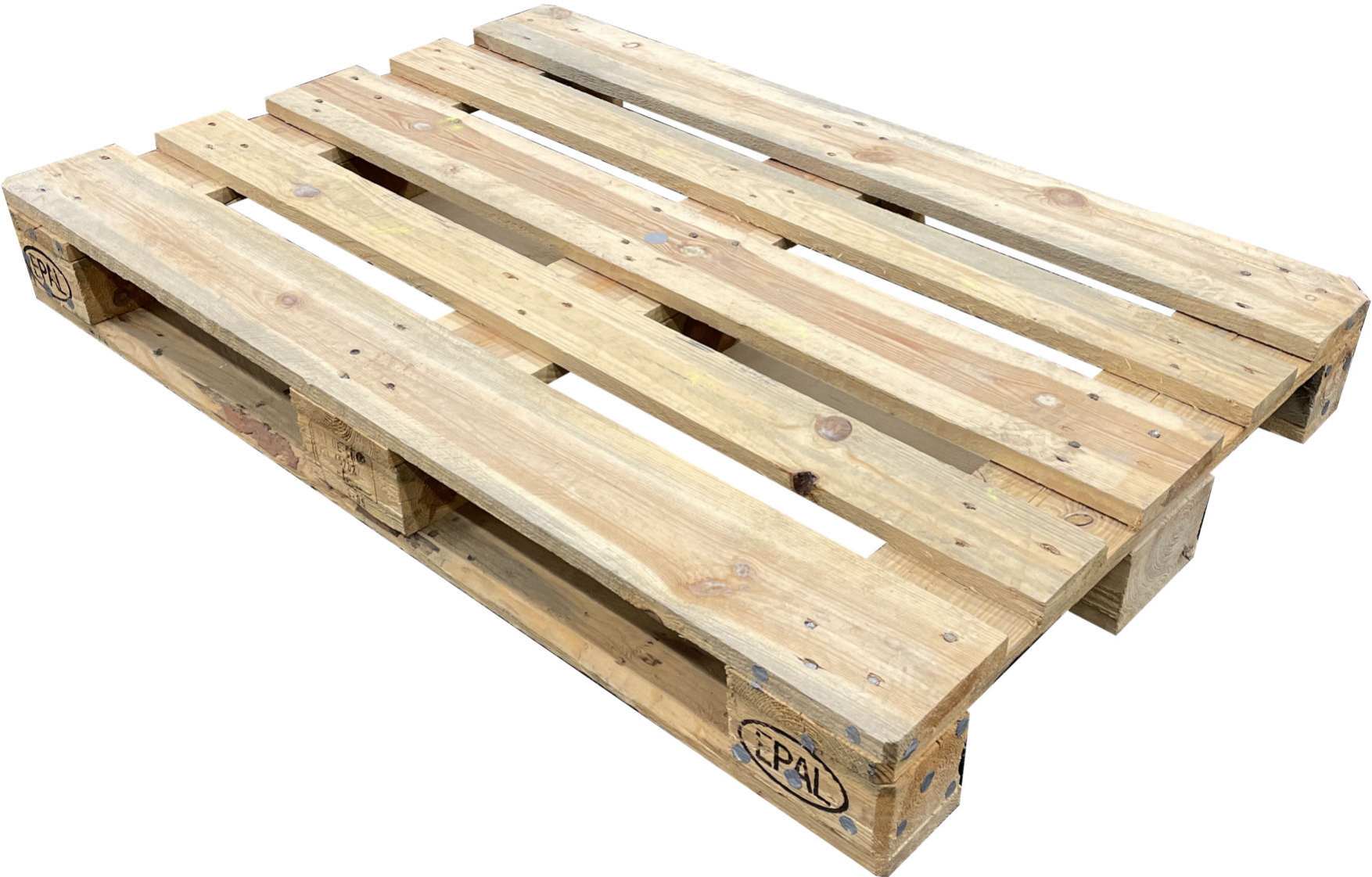}
        \caption{}
        \label{fig:pallet}
    \end{subfigure}
    \hfill
    \begin{subfigure}[b]{0.19\textwidth}
        \includegraphics[width=\textwidth]{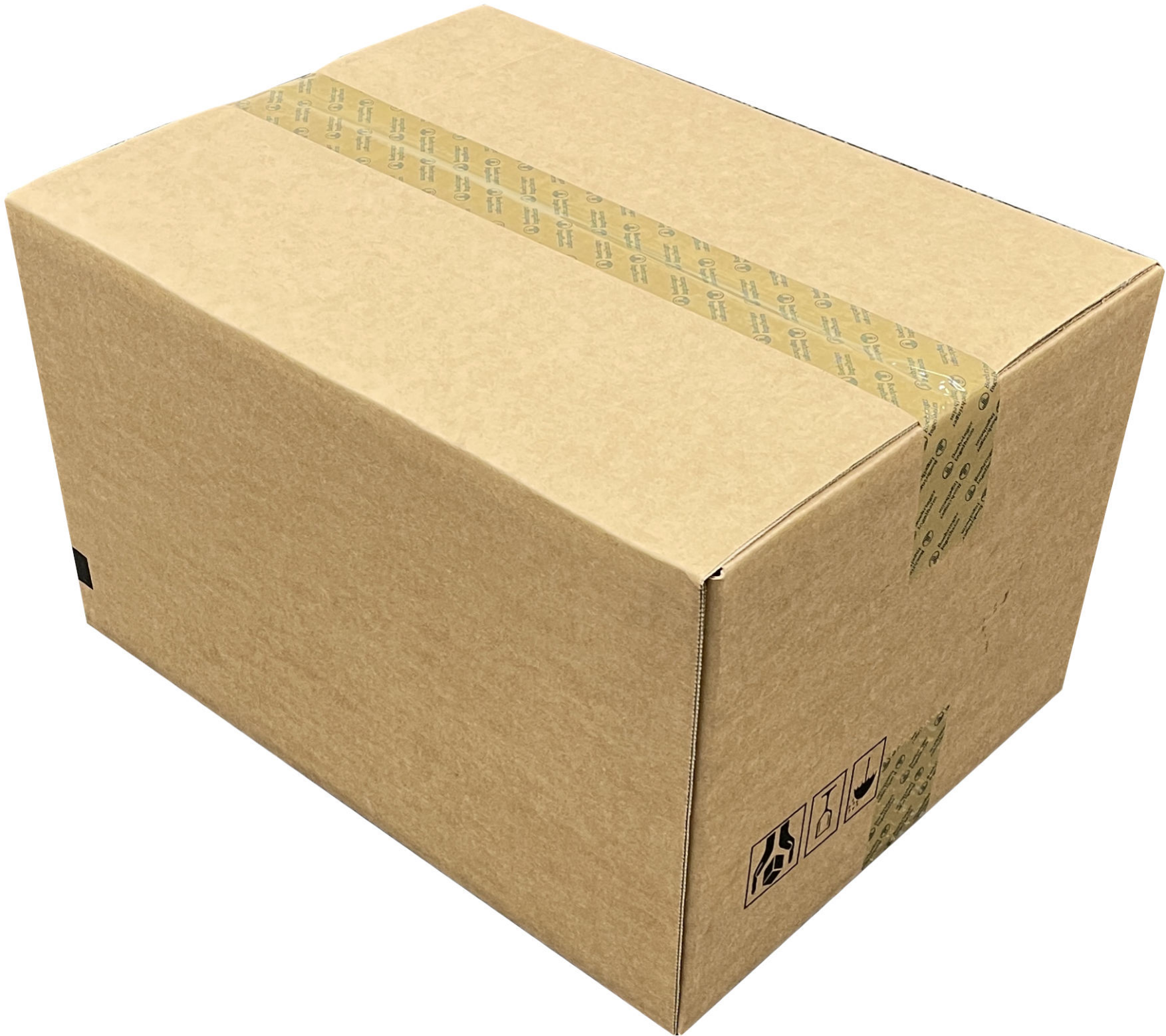}
        \caption{}
        \label{fig:cardboard_box}
    \end{subfigure}
    \hfill
    \begin{subfigure}[b]{0.19\textwidth}
        \includegraphics[width=\textwidth]{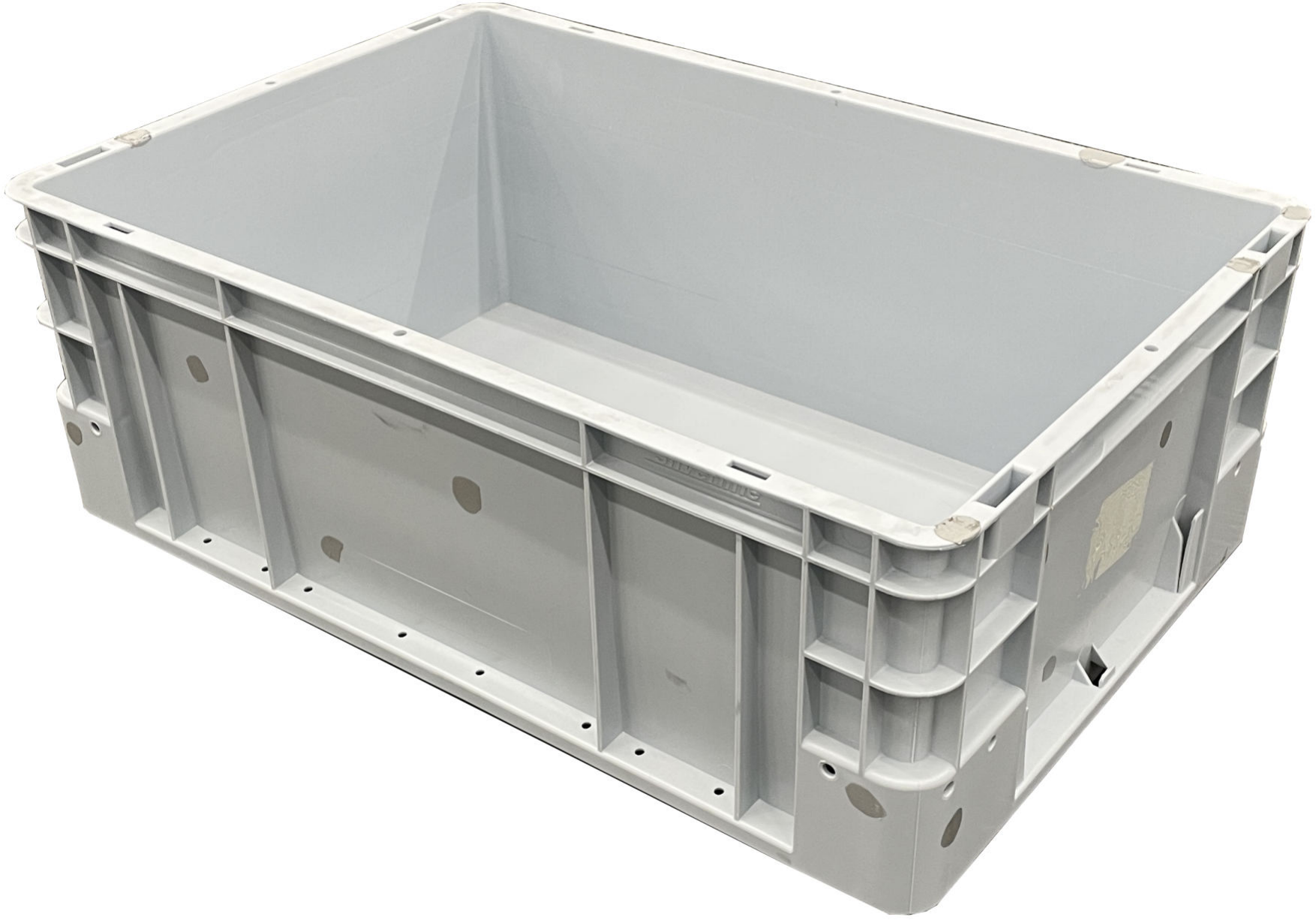}
        \caption{}
        \label{fig:klt}
    \end{subfigure}
    \hfill
    \begin{subfigure}[b]{0.19\textwidth}
        \includegraphics[width=\textwidth]{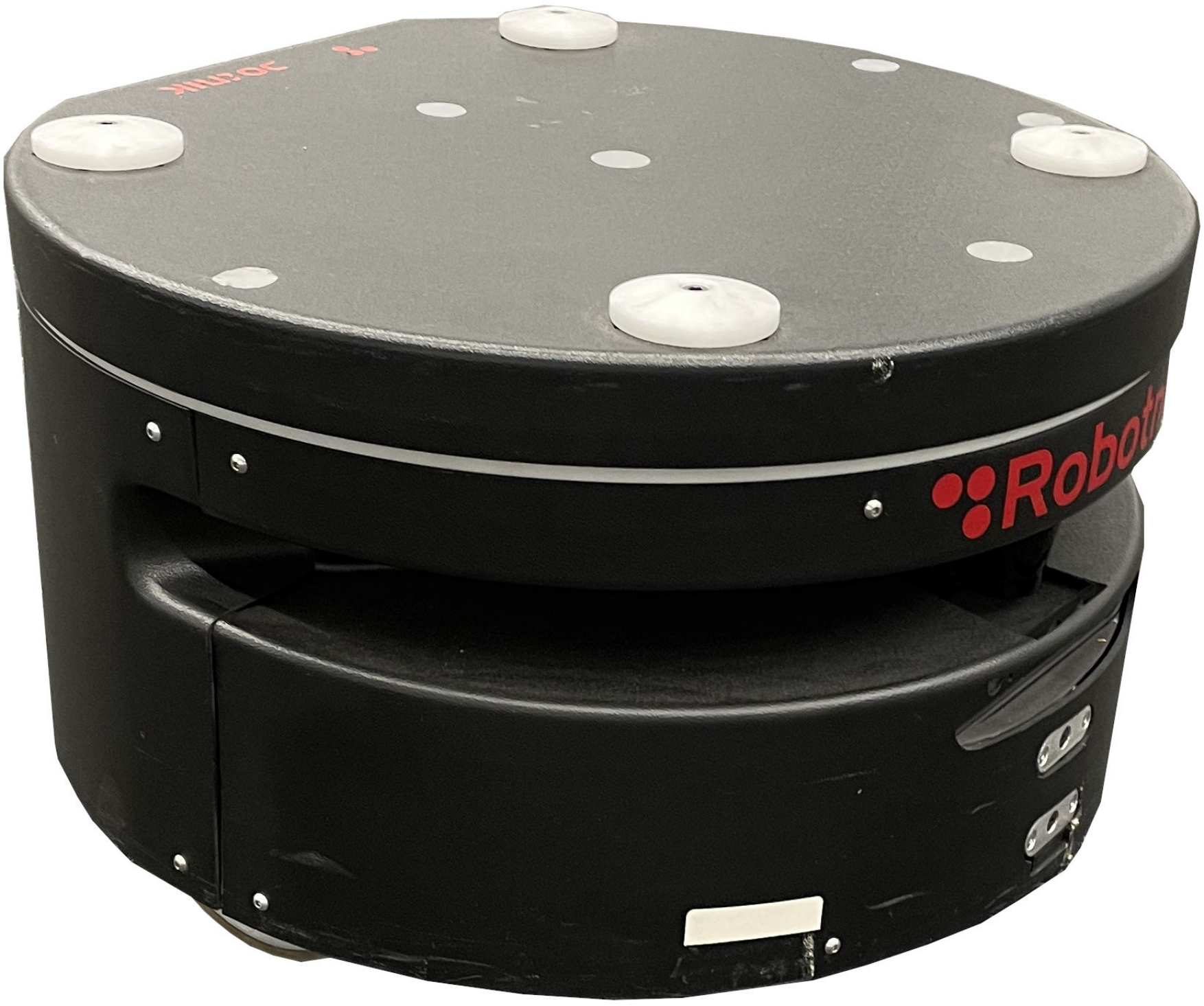}
        \caption{}
        \label{fig:robotnik}
    \end{subfigure}
        \hfill
    \begin{subfigure}[b]{0.19\textwidth}
        \includegraphics[width=\textwidth]{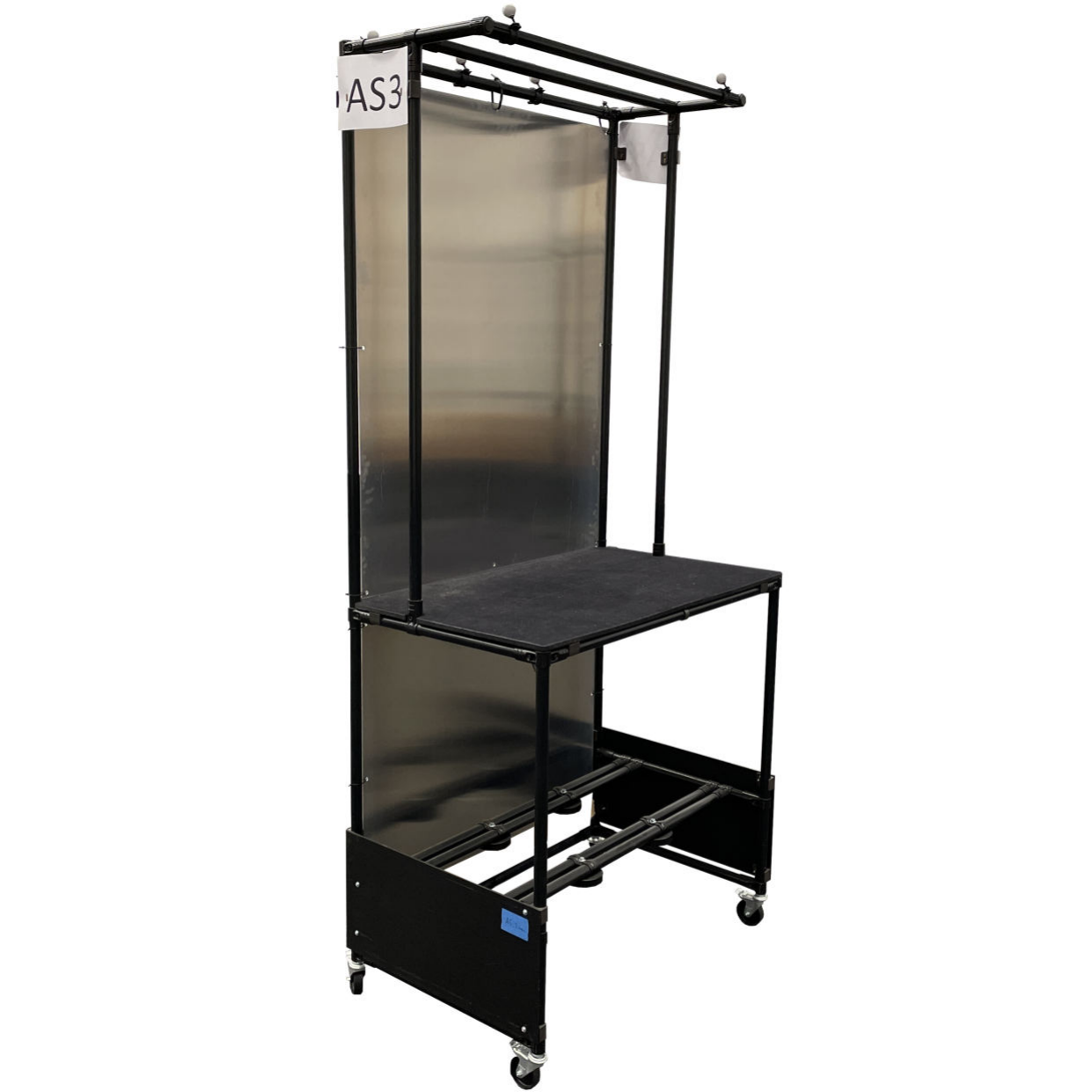}
        \caption{}
        \label{fig:workstation}
    \end{subfigure}
    \caption{The objects used in Multi-log are a) pallets, b) cardboard boxes, c) small load carriers, d) mobile robots, and e) movable workstations.}
    \label{fig:multi-log_objs}
\end{figure}

The dataset is comprised of five object classes, namely pallets, cardboard boxes, small load carriers, mobile robots, and movable industrial workstations (see \cref{fig:multi-log_objs}). The objects have a total of nine physical instances that differ in color and texture.
Each object is marked in the motion capture system for tracking purposes. 3D frame axes are attached virtually to the volumetric center of each object at a pre-measured location which is consistent with the origin location of the 3D models collected. We separate the data collection stage from the annotation stage to preserve the raw data and to increase the recording rate by isolating the computationally demanding annotation stage. 

The collected images are formatted in a scene structure, in accordance with the BOP format \cite{ferrari_bop_2018}. Our  format differs, however, in that we define a scene as being a \emph{snap} of the current environment through our eight-camera recording setup. Thus, each scene could, at most, contain eight images. 
The scene includes multiple objects with different poses, but each image is associated with the poses of all possible objects in the scene, even if they are not visible in the camera's perspective. During the annotation stage (see \cref{sec:pipeline}), object projections outside the image plane of each camera are filtered out to obtain only relevant objects for each camera, ensuring accurate and relevant annotations.
In order for our dataset to resemble the BOP format as closely as possible, we provide \emph{mock} depth images that are our aggregated masks with a fixed distance from the camera. Thus, we assume that objects do not occlude one another significantly in the collected dataset. We deem this assumption to be valid due to the elevated vantage point of the cameras and the large area of operation for objects in the scene. Our dataset is publicly available.


\begin{figure}[t!]
    \centering
    \includegraphics[width=1.0\linewidth]{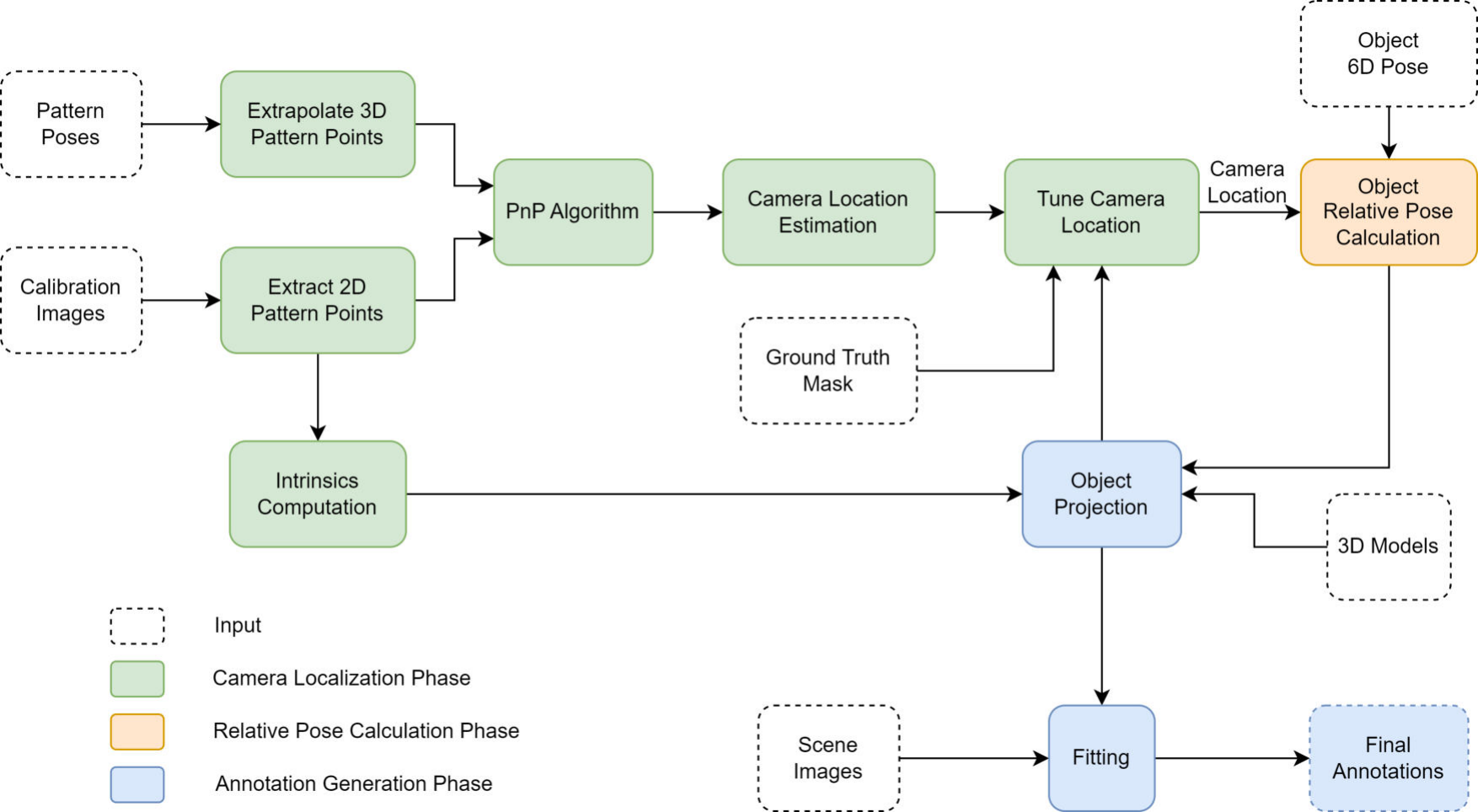}
    \caption{The automated annotation pipeline developed for this contribution.}
    \label{fig:pipeline}
\end{figure}

\section{Automated Annotation Pipeline} \label{sec:pipeline}

We devise an automated annotation pipeline to simplify the annotation process for large datasets. The pipeline has three phases, including unifying the reference frames of the RGB camera and motion capture systems, computing relative transformations, and generating annotations. An overview of the pipeline is provided in the figure mentioned as \cref{fig:pipeline}. In the upcoming sections, we discuss the details of each of the aforementioned annotation pipeline phases. Our source code is publicly available \footnote{https://anonymous.4open.science/r/bop\_toolkit-6F86}. 


\subsection{RGB Camera Localization} \label{sec:cam_loc}



In industrial environments, cameras are often mounted for monitoring purposes, and localizing them with respect to a fixed reference frame is beneficial for object localization due to increased system control and flexibility. Finding the relative poses of objects requires the camera pose since objects are localized only with respect to the motion capture system's reference frame. Manually measuring camera localization is challenging due to accumulated errors, and is dependent on the camera hardware's design and lens, thus limiting generalizability.

\begin{figure}[h]
    \centering
    \begin{subfigure}[b]{0.49\textwidth}
        \includegraphics[width=\textwidth]{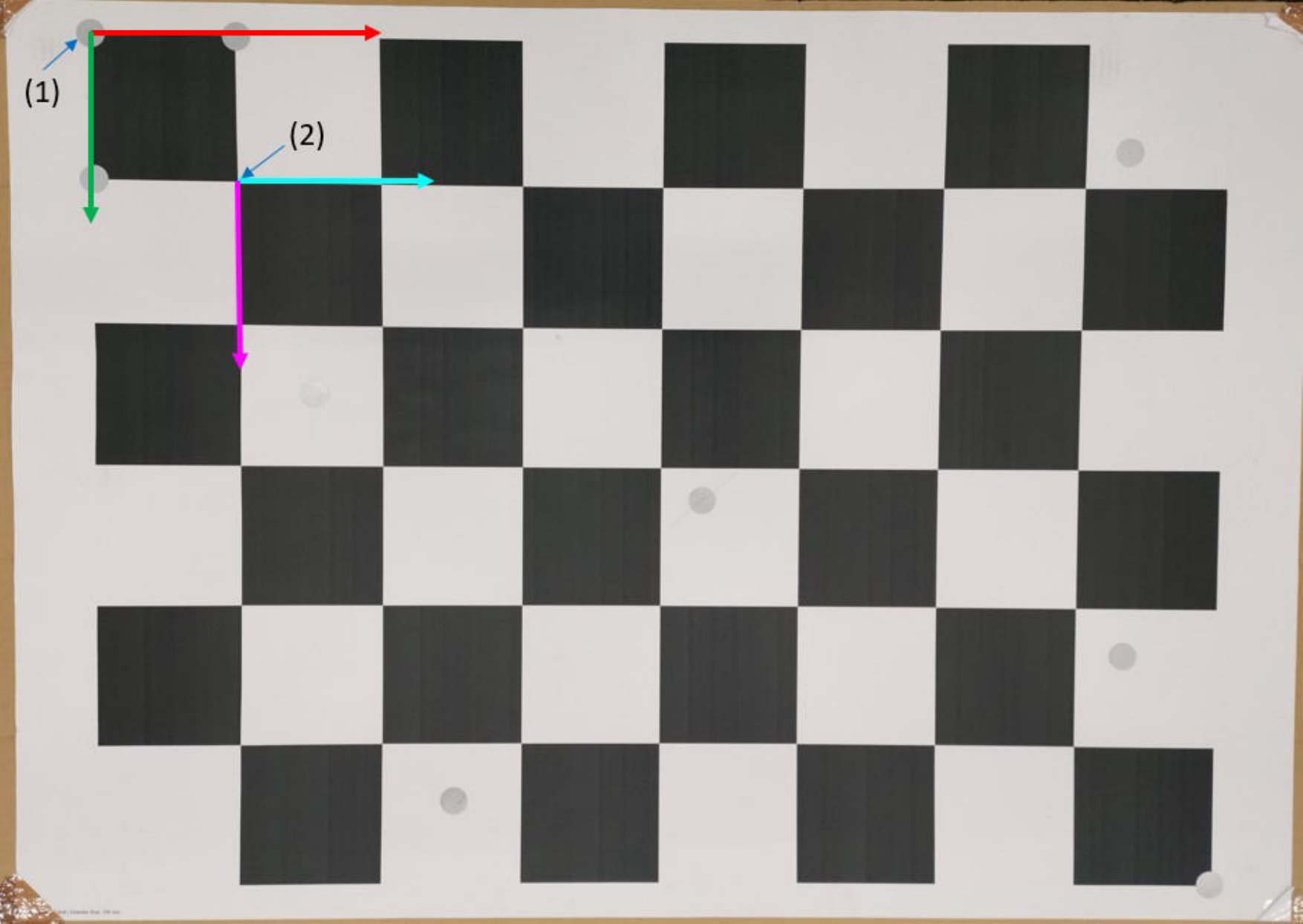}
        \caption{}
        \label{fig_part_a}
    \end{subfigure}
    \begin{subfigure}[b]{0.49\textwidth}
        \includegraphics[width=\textwidth]{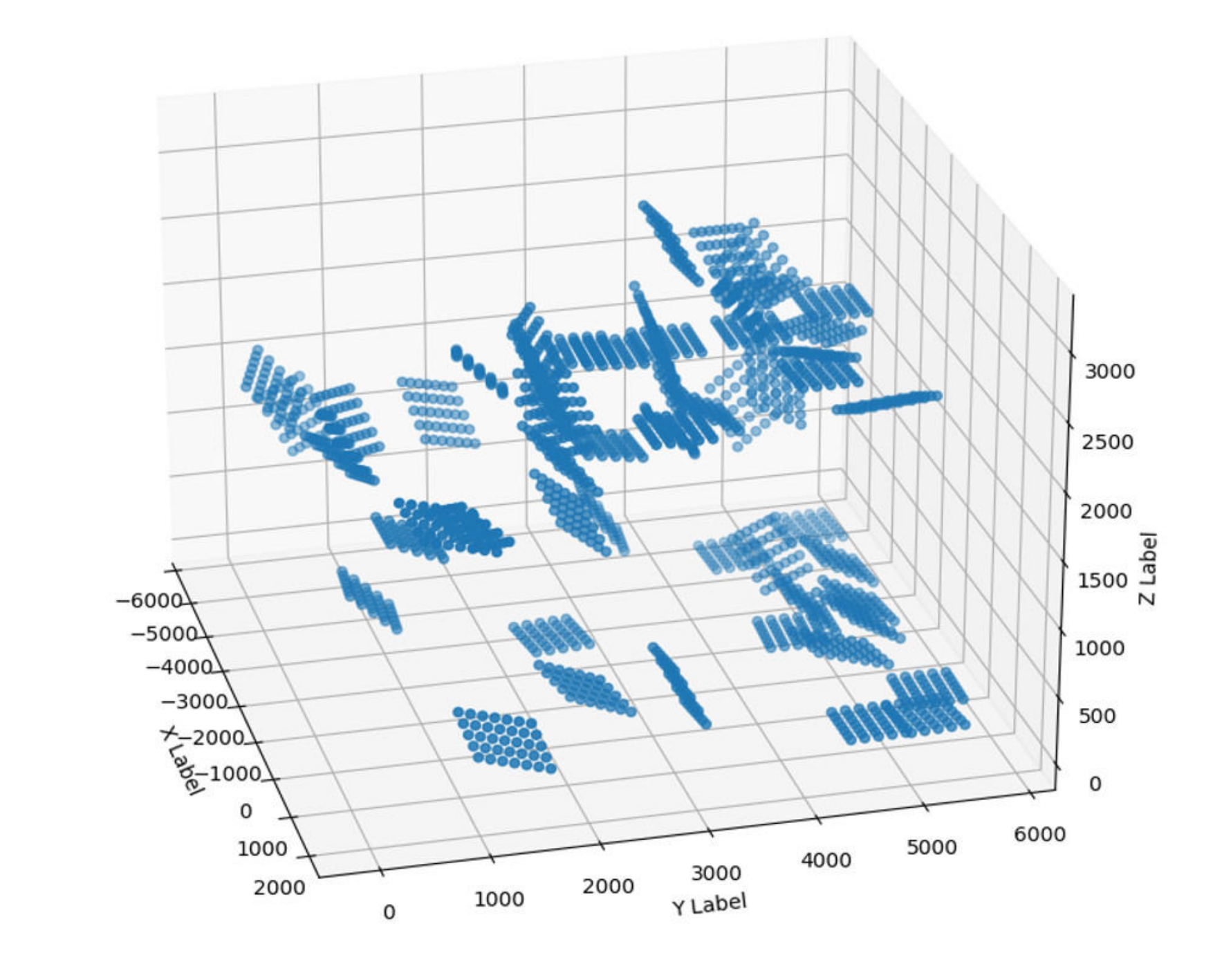}
        \caption{}
        \label{fig_part_b}
    \end{subfigure}
    \caption{(a) A tracked checkerboard pattern was used to obtain camera intrinsic parameters for each camera and to unify the reference frame for the motion capture and the RGB camera systems. (1) shows the object origin in the motion capture system's global reference frame. Red and green axes correspond to the X and Y directions, respectively, of the virtually attached frame. (2) shows the X and Y directions (in cyan and magenta, respectively) of the pixels on the image originating from the first intersection. (b) The visualization of checkerboard pattern positions used during the camera localization process. The camera is situated in the upper right-hand corner.}
    \label{fig:checkerboard_and_cam_loc}
\end{figure}

To deal with these issues, we devised a custom method to localize the cameras in an adaptable manner. Our approach uses a tracked checkerboard pattern, as shown in \cref{fig:checkerboard_and_cam_loc}, that is detectable by both systems. The pattern is large enough ($841$ mm~×~$1189$ mm) to increase the detectability of its intersection points by the RGB cameras in a manner similar to the commonly used Zhang's method \cite{zhang_flexible_2000} for intrinsic camera calibration. For tracking using the motion capture system, retro-reflective markers have been added randomly to the checkerboard, except for the precisely-located markers defining the outer boundary points of the upper left checkerboard box. The motion capture system provides the location of the checkerboard pattern with respect to the upper left corner where the checkerboard's virtual frame is attached, as illustrated in \cref{fig:checkerboard_and_cam_loc}. Using the known size of the pattern and the homogeneous transformations between both systems and the patterns, the location for each RGB camera can be obtained. 

We formulate the problem of localizing the cameras in space as an object pose estimation problem, where the pose of a camera is to be obtained with respect to the reference frame of the motion capture system using a tracked pattern as an intermediate tool. To this end, we make use of classical object pose estimation algorithms, namely the Perspective-n-point (PnP) algorithm \cite{lepetit_epnp_2009}. 

The camera localization process
starts by collecting checkerboard images in a manner similar to that used in standard camera calibration procedures, such as with Zhang's method \cite{zhang_flexible_2000}. The pixel intersection points on the checkerboard are extracted from the collected images and concatenated. Simultaneously, the 6D pose of the checkerboard is retrieved by the motion capture system for each image capture. The aim is to retrieve 2D points on the checkerboard and their corresponding 3D points in space, both of which would be used by the PnP algorithm to obtain the relative pose between the camera and the checkerboard pattern. However, as shown in \cref{fig_part_a}, an offset exists between the first checkerboard intersection point and its virtual origin in space. The offset is corrected via a static homogeneous transformation:

\begin{equation} \label{eq:checkerboard_offset}
    H_{mc}^{ch_{inter}} = H_{mc}^{ch_{origin}}H_{ch_{origin}}^{ch_{inter}}
\end{equation}

Where $H_{mc}^{ch_{inter}}$ represents the homogeneous transformation of the checkerboard pattern's first intersection point with respect to the motion capture system reference frame. $H_{mc}^{ch_{origin}}$ is the homogeneous transformation between the checkerboard pattern's virtual origin, located in the top left corner, and the motion capture system. Finally, $H_{ch_{origin}}^{ch_{inter}}$ is a static homogeneous transformation with a translation vector obtained from the dimensions of the checkerboard pattern and an identity orientation.

Using \cref{eq:checkerboard_offset}, the 3D vector representing the corresponding point in space to the first intersection point of the checkerboard's pattern could be obtained. The remaining intersection points of the pattern are derived using a homogeneous static transformation at each extrapolated point, similar to that in \cref{eq:checkerboard_offset}, with a rotational component that is equal to the board's orientation. \cref{fig_part_b} shows the resulting extrapolated point for each of the pattern's poses.
The concatenated set of 2D image points and their corresponding 3D points are then passed to the PnP algorithm. Since all 3D points are in the motion capture system space, the resulting output of the solvePnP algorithm is the camera's pose with respect to the motion capture system's reference frame. This also unifies the reference frames of both systems.

The main aim of the aforementioned steps is to find a roughly accurate camera location. The initial camera location is subpar due to errors emerging from the detection of the 2D intersections of the patterns, as well as the extrapolation of their corresponding 3D points in space. Thus, we apply a further tuning step to compensate for the errors in the localization process.


 The tuning process is performed by manually creating binary masks for objects of interest in sample images. A range of offsets is defined as the search space of possible camera poses with respect to the initially retrieved pose. The projection of the 3D models of corresponding objects was then obtained at their calculated relative ground truth at each entry of the pre-defined camera pose range. The intersection between the binary mask and the projected object mask is deduced for each image in the search space. Poses for intersections surpassing a pre-defined threshold are accepted as the final poses of the camera under investigation. The process is then repeated for all cameras. The tuning is only done in an initial phase, as shown in \cref{fig:pipeline}, until the overlap of the projected mask with the ground truth masks surpassed a pre-defined threshold. Once accurate poses for the cameras are ensured, the tuning is halted and the final camera locations are used to calculate the relative object poses.

\subsection{Calculation of Relative Poses}
Datasets collected in a manner similar to our custom dataset, as discussed in \cref{sec:dataset}, contain global poses of the objects of interest and images thereof.
The motion capture system provides the poses of the objects with respect to its global reference frame. However, pose estimation is the problem of finding the pose of the object with respect to the camera frame. Such a relative transformation is calculated as a result of a transformation chain between the pre-obtained locations of the cameras and the global location of the object. The calculation of the relative transformation is applied to all objects of interest, in all obtained images, in an offline manner using the obtained camera pose, as discussed in \cref{sec:cam_loc}, and the object 6D pose. The transformation chain can be described as follows:

\begin{equation}
    H_{obj}^{cam} = (H_{cam}^{mc})^{-1}H_{obj}^{mc}
\end{equation}

Where $H_{obj}^{cam}$ represents the homogeneous transformation of the object with respect to the camera (relative transformation). 
$H_{cam}^{mc}$ and $H_{obj}^{mc}$ represent the homogeneous transformations between the camera and motion capture system, and between the object and motion capture system, respectively.

The calculation of the relative poses is part of the camera location tuning procedure, as illustrated in \cref{fig:pipeline}, and it enables the projection of object models onto sample images in order to match them with ground truth masks, as discussed in \cref{sec:cam_loc}. It is worth noting that the calculation of the relative poses of the objects of interest in our custom dataset defaults to using the final camera locations after the camera tuning step is carried out. 

\subsection{Annotation Generation}
The aim of the annotation generation phase is three-fold: First, to format the collected data in a scene structure, then to generate annotations such as masks, bounding boxes, etc., and finally to filter invalid images. The relative poses, obtained in the previous phase, are used as the ground truth poses for the objects captured in the scene. The 3D models of the objects of interest are then rendered at their respective ground truth locations, using VisPy visualization library \cite{luke_campagnola_2022_5974509} as part of the BOP toolkit \cite{ferrari_bop_2018}, and then projected on the image plane using the camera parameters. The projection of the 3D points to pixel locations is accomplished using the well-known projection matrix \cite{hartley_multiple_2004}:
\begin{equation}
    x = PX 
\end{equation}
where $X$ is a $4$~×~$1$ vector of a point location in 3D space, $x$ is a $3$~×~$1$ vector of pixel locations, and $P$ is the projection matrix defined as:
\begin{equation}
        P = K[R|t] = KR[I|R^Tt]
\end{equation}
where $K$ is the $3$~×~$3$ camera matrix describing the intrinsic parameters of the camera. $R$ is the $3$~×~$3$ rotation matrix and $t$ is the $3$~×~$1$ translation vector.
The projected models are then aggregated to get all object masks for each input image. Projected masks that reside outside an image are filtered. The pipeline then fits each of the projected object masks with a 2D bounding box, as shown in \cref{fig:process} to form our final annotations. 


\section{Results}
\begin{figure}[t!]
    \centering
    \begin{subfigure}{0.24\linewidth}
        \includegraphics[width=\linewidth]{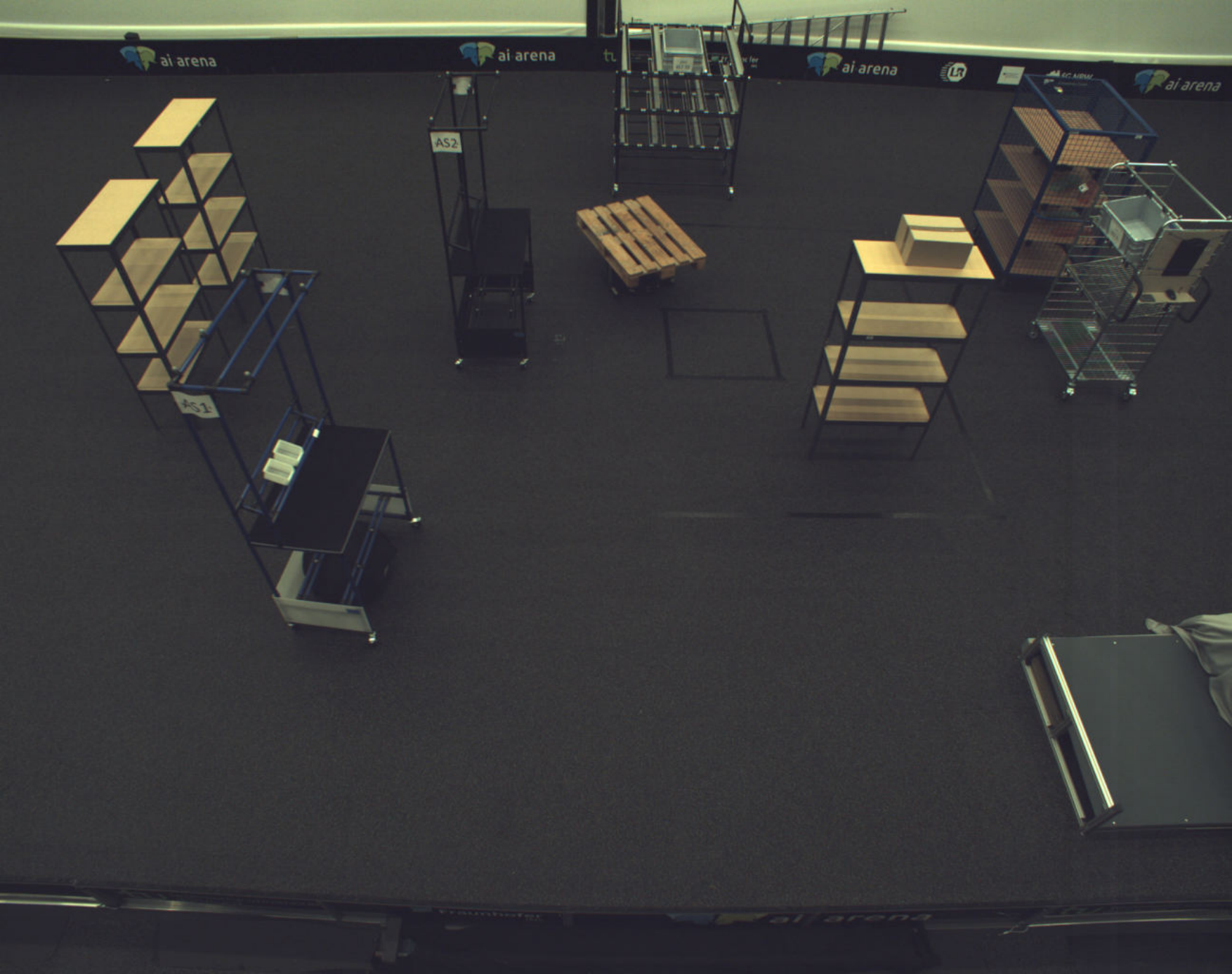}
        \caption{}
        \label{fig:process_a}
    \end{subfigure}
    \begin{subfigure}{0.24\linewidth}
        \includegraphics[width=\linewidth]{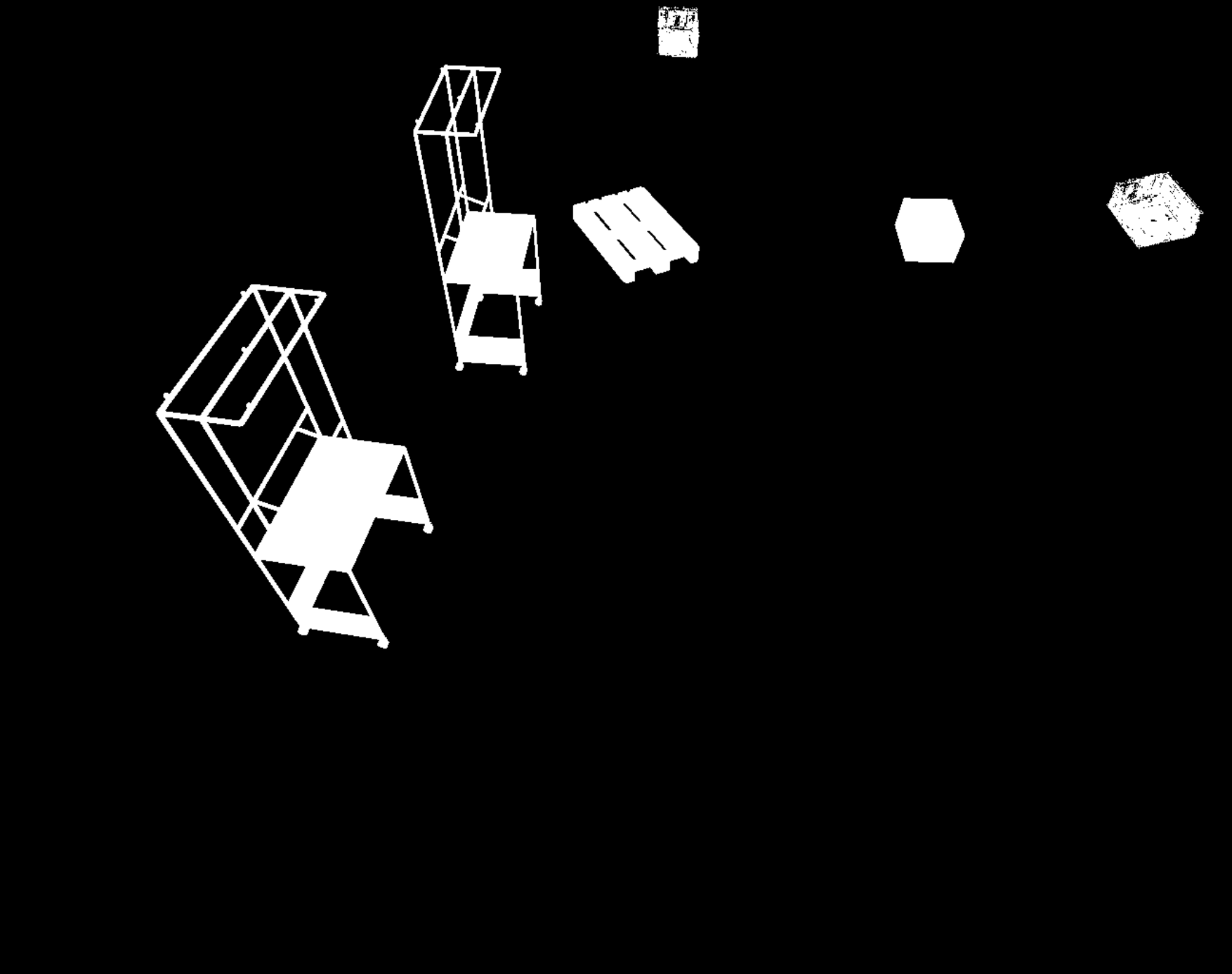}
        \caption{}
        \label{fig:process_b}
    \end{subfigure}
    \begin{subfigure}{0.24\linewidth}
        \includegraphics[width=\linewidth]{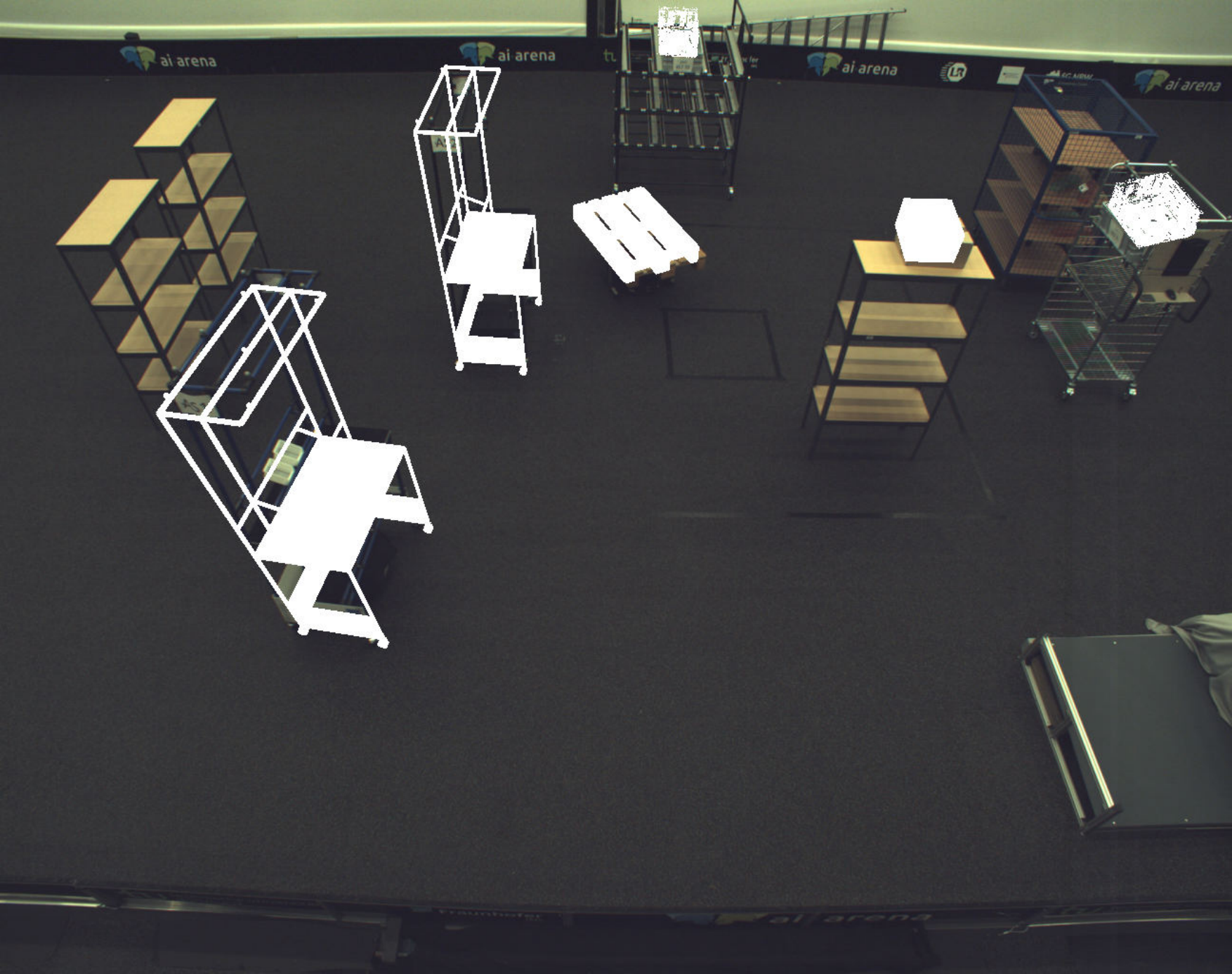}
        \caption{}
        \label{fig:process_c}
    \end{subfigure}
    \begin{subfigure}{0.24\linewidth}
        \includegraphics[width=\linewidth]{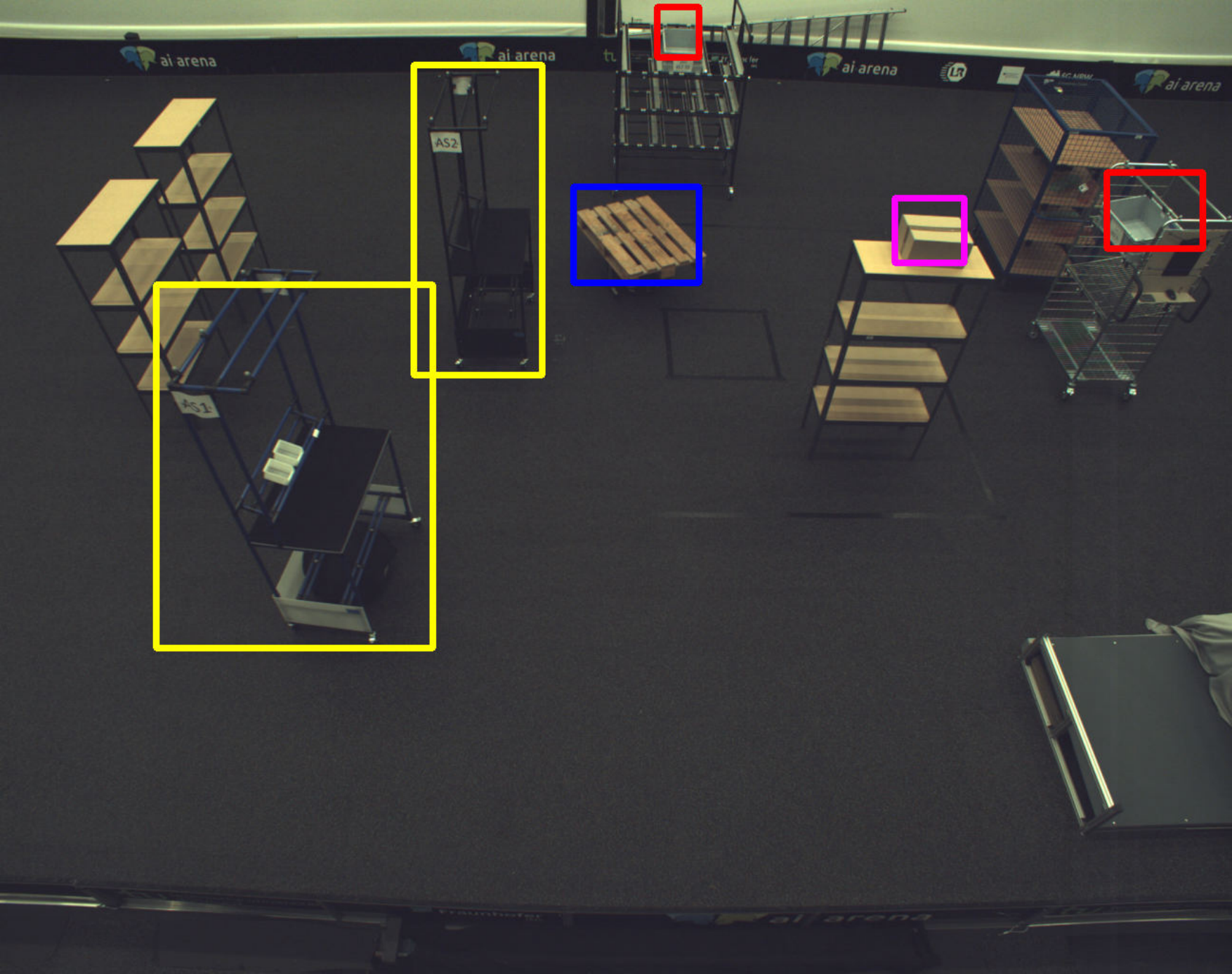}
        \caption{}
        \label{fig:proces_d}
    \end{subfigure}
    \caption{(a) Sample image with various tracked objects, (b) mask of all visible tracked objects which is used only initially in the camera localization phase, (c) overlay of objects' 3D models at the calculated poses, (d) final object annotation derived from object masks.}
    \label{fig:process}
\end{figure}

The dataset presented in this work consists of $6,136$ images with $26,500$ different object instances in total.
The total amount of time spent during annotation of all images is about $13.9$ hours.
This results in an average of $1.9$ seconds spent on the annotation of each object instance.
In comparison to the time spent on manual annotation, our pipeline results in a substantial increase in annotation speed.
A visualization of the results of the individual phases is shown in \cref{fig:process}.

\begin{table}[h!]
    \centering
    \caption{Dataset statistics per camera.}
    \setlength{\tabcolsep}{5pt} 
    \renewcommand{\arraystretch}{1.2} 
    \begin{tabular}{r| r r}
        \textbf{Sequence}       &   Scenario I      &   Scenario II    \\\hline
        Number of instances            &   16,678          &   9,804           \\
        Number of frames               &   3,920           &   2,216 \\
        Annotation time [min]   &   525             &   307    \\
    \end{tabular}
    \label{tab:stat_cam_1}
\end{table}

Scenarios I and II recorded a total of about $26,500$ object instances.
These object instances are composed of the five objects selected for this dataset.
The dataset images were recorded at half the available resolution by the camera system, resulting in $1296$~×~$1024$ images. The reduced resolution enabled stream capturing at a higher frame rate of about $5$ FPS.
Individual statistics per scenario are shown in \cref{tab:stat_cam_1}. 

\begin{figure}[h!]
    \centering
    \begin{subfigure}{0.49\linewidth}
        \includegraphics[width=\linewidth]{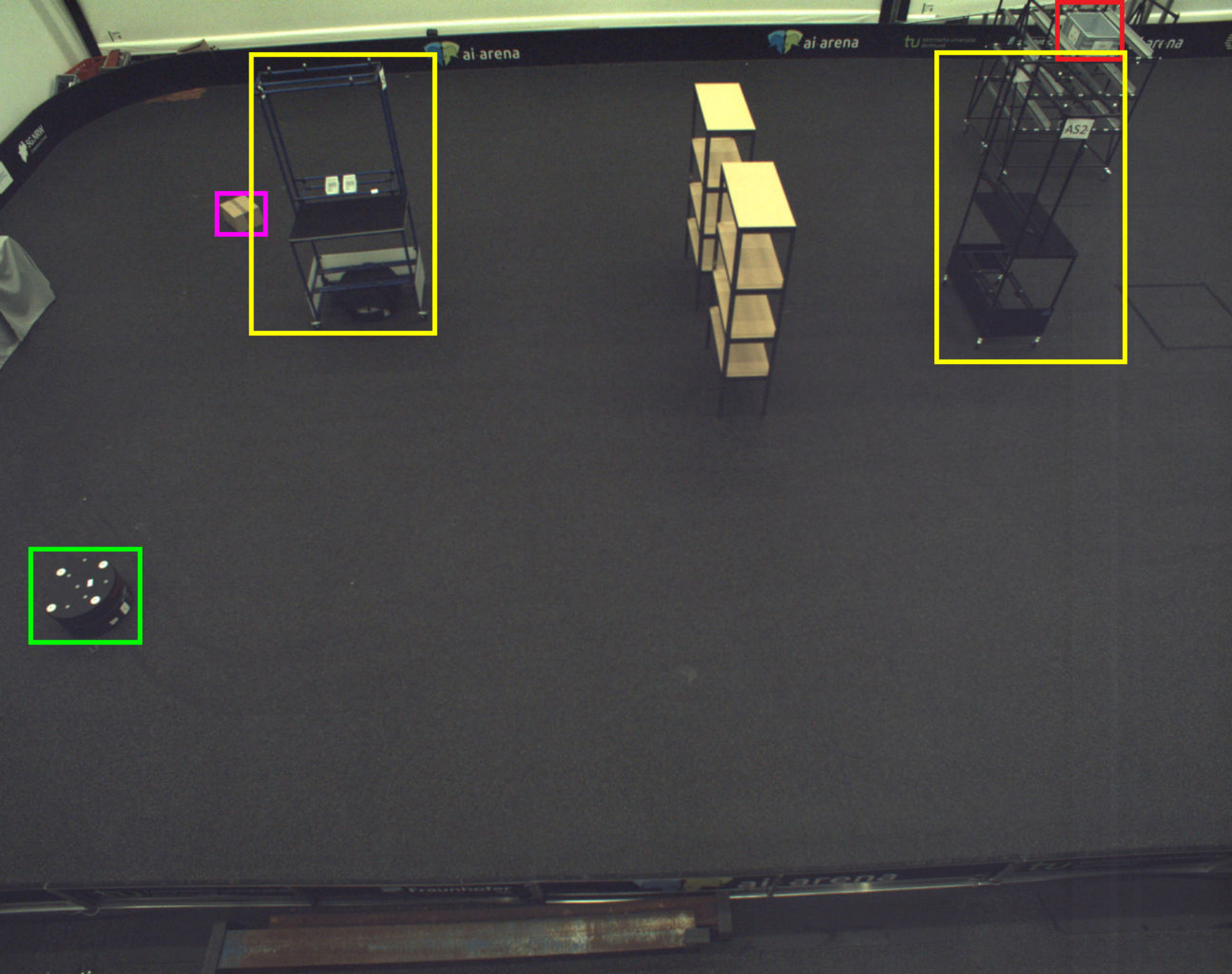}
    \end{subfigure}
    \begin{subfigure}{0.49\linewidth}
        \includegraphics[width=\linewidth]{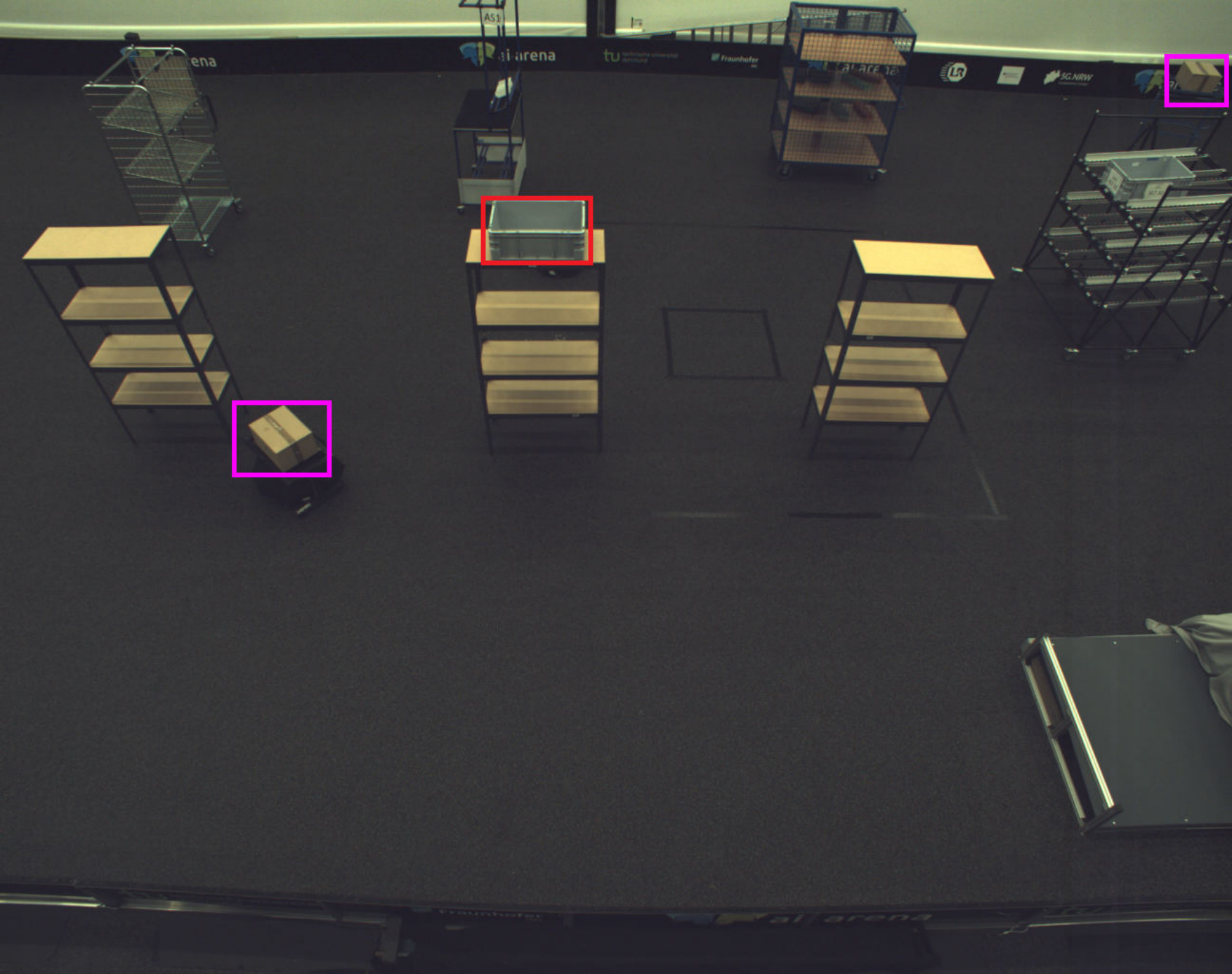}
    \end{subfigure}
    \caption{Visualization of the annotation pipeline results on the Multi-log dataset.}
    \label{fig:results}
\end{figure}

Of the total object instance captured over the two scenarios, the small load carrier is overly represented due to the utilization of multiple carrier instances per scenario.
In total, $7,363$ instances of the small load carrier were captured.
This is closely followed by $6,587$ instances of pallets and $6,403$ instances of cardboard boxes.
Workstation and robot instances were captured the least with $3,768$ and $2,361$ instances respectively. 
An excerpt of the dataset with all object instances is shown in \cref{fig:results}



\section{Conclusion}
In this work, we present a pipeline to automatically annotate monocular images using ground truth poses of objects of interest. As part of the pipeline, we also devise a methodology to localize freely-mounted cameras in space. We test our pipeline on a custom dataset collected from an industrial-like setting. The final results show the efficiency of our annotation pipeline. Our approach is generalizable to settings where 6D object poses are readily available with respect to a fixed reference frame. We would like to extend the testing of our pipeline to larger datasets to validate the scaling of our methodology. Also, using the annotated data, we would like to train baseline architectures for object pose estimation and object detection either from scratch or as part of a transfer learning pipeline.

\subsubsection{Acknowledgements.} 
This work received funding from the German Federal Ministry of Education and Research (BMBF) in the course of the Lamarr Institute for Machine Learning and Artificial Intelligence (LAMARR23B).

%
%
%
\bibliographystyle{splncs04}
\bibliography{Main}

\end{document}